\documentclass[9pt]{article}
\usepackage[numbers]{natbib}

\PassOptionsToPackage{numbers}{natbib}
\usepackage[preprint]{neurips_2024}
\usepackage[utf8]{inputenc} 
\usepackage[T1]{fontenc}    
\usepackage{url}            
\usepackage{booktabs}       
\usepackage{array}          
\newcolumntype{C}[1]{>{\centering\arraybackslash}p{#1}}
\usepackage{amsfonts}       
\usepackage{nicefrac}       
\usepackage{microtype}      
\usepackage{xcolor}         
\usepackage{enumitem}
\usepackage{amsmath}
\usepackage{graphicx}
\usepackage{placeins}
\usepackage{float}
\usepackage{comment}
\usepackage{mathtools}
\newcommand{\thead}[1]{\textbf{\shortstack{#1}}}
\usepackage{tikz}
\usetikzlibrary{decorations.text, arrows.meta, positioning, shadows, calc, shapes.symbols}
\definecolor{nvidiagreen}{RGB}{76,175,80}
\colorlet{nvidiaLight}{nvidiagreen!25!white}   
\colorlet{nvidiaMed}{nvidiagreen!60!white}     
\colorlet{nvidiaDark}{nvidiagreen!85!black}    
\newcommand{\editgreen}[1]{#1}

\usepackage[colorlinks=true, urlcolor=blue, citecolor=blue, linkcolor=blue]{hyperref}
\usepackage{multirow}
\usepackage{xspace}
\usepackage[framemethod=TikZ]{mdframed}
\mdfsetup{skipabove=0.5em, skipbelow=0.5em, innerleftmargin=8pt, innerrightmargin=8pt, innertopmargin=6pt, innerbottommargin=6pt}

\makeatletter
\renewcommand{\paragraph}{%
  \@startsection{paragraph}{4}{\z@}%
                {0.8ex \@plus 0.3ex \@minus 0.1ex}%
                {-1em}%
                {\normalsize\bf}%
}
\makeatother

\title{Scaling Laws for Task-Specific LLM Distillation}

\author{%
  \textbf{Lavinia Ghita}\thanks{Equal contribution. Correspondence to: \texttt{lghita@nvidia.com}, \texttt{dhruvd@nvidia.com}.}~
  \And
  \textbf{Dhruv Desai}\footnotemark[1]~
  \And
  \textbf{Ioana Boier}\\ \\
  \hspace{-8.4cm} NVIDIA
}

\begin{document}

\maketitle

\begin{abstract}

Large Language Models (LLMs) achieve strong performance across a growing range of domains, yet their scale poses deployment challenges in applications where latency and cost constraints are critical. This paper derives empirical scaling laws for domain-specific LLM compression, quantifying how in-domain and general-knowledge performance scale with dataset size, compression ratio, supervision format, and iterative pruning schedule. Using quantitative finance as our application domain, we compare logit-based and LoRA-based distillation \editgreen{on structurally pruned students}, introducing a blended chain-of-thought supervision loss that stabilizes KL-divergence distillation over reasoning traces. In-domain task quality degrades predictably under compression while general-knowledge benchmarks collapse well before the same point; supervision format is the key driver of this tradeoff, with chain-of-thought supervision actively recovering general knowledge that pruning erases. We release the headline dataset \href{https://huggingface.co/datasets/nvidia/FinHeadlineMix}{\texttt{FinHeadlineMix}}, scaling-law results, and practical recommendations to provide a reusable framework for domain-specific compression decisions.

\end{abstract}

\section{Introduction}

We study knowledge distillation \editgreen{(KD)} for compressing LLMs in a \emph{domain- and task-specific} setting, quantifying the tradeoff between preserving in-domain task performance and retaining broad general-knowledge capabilities. This tradeoff is of direct interest to practitioners who need smaller, deployable models under latency and cost constraints. Concretely, the domain-specific task we use is a 35-class event classification problem over financial news headlines: a narrow, fixed-taxonomy mapping representative of production workloads. We compare multiple distillation methods and supervision formats (label-only versus chain-of-thought), apply iterative structural pruning and distillation, and derive empirical scaling laws across dataset size, compression ratio, supervision format, and pruning schedule. We provide practical guidance for choosing methods and compression levels. We find that supervision format governs this tradeoff: blended chain-of-thought supervision not only improves data efficiency but actively recovers general knowledge lost to pruning, while label-only methods either preserve the pruned level (LoRA) or degrade it further (direct-label KD). Iterative pruning and distillation compress the teacher to 16\% of its parameters while retaining meaningful task quality; self-distillation baselines show that the training process itself accounts for roughly twice the performance gap introduced by compression.

We ground our domain-specific experiments in quantitative finance, where these tradeoffs are acute. Synthetic annotation with LLMs scales the creation of labeled data \cite{Nvidia:2024aa}, letting practitioners iterate quickly on data sources and signal hypotheses. Production systems in finance are latency- and cost-sensitive, making compression essential and providing a natural testbed for studying it under realistic deployment constraints. Our methodology is domain-agnostic. LLM-based annotation shifts the bottleneck from labeling to deployment, motivating the compression study that follows.

We focus on compression via fewer parameters: pruning \cite{NIPS1989_6c9882bb, Muralidharan:2024aa} and knowledge distillation \cite{Hinton:2015aa}. Recent work \cite{Bercovich:2025aa, DeepSeek-AI:2025aa, Muralidharan:2024aa} shows these techniques can significantly reduce size while largely preserving accuracy. Quantization, sparsity, and neural architecture search (NAS) are outside our scope. We compare LoRA-based \cite{Hu:2021aa, schulman2025lora} and logit-based \cite{Muralidharan:2024aa} distillation, each with single-label and chain-of-thought (CoT)~\cite{Wei:2022aa} outputs. To stabilize KL-divergence distillation over reasoning traces, we introduce a blended supervision loss that independently weights label and CoT tokens (Section~\ref{sec:blended_loss}). To our knowledge, this is the first systematic comparison of these methods on decoder-only LLM students at multi-billion parameter scale.

The rest of the paper is structured as follows: Sections~\ref{sec:related_work}--\ref{sec:methods} present related work, data, and methodology; Section~\ref{sec:experiments} presents the experimental results and scaling laws; Section~\ref{sec:conclusion} covers the conclusions and a broader discussion aimed at practitioners. The Appendix gives implementation details, dataset description, and supplementary results. We release the dataset used in our study.

\section{Related Work}
\label{sec:related_work}

Knowledge distillation was introduced as a compression method in \cite{Bucilua:2006}, and later formalized in its modern form by \cite{Hinton:2015aa}. This foundational approach distinguishes between hard distillation, where the student learns from the teacher's argmax labels, and soft distillation, where the student matches the teacher's full output distribution via softened logits.

An extension using chain-of-thought (CoT) \cite{Wei:2022aa} as supervision is discussed in \cite{Hsieh:2023aa}, showing that richer teacher supervision can match label-only distillation performance with significantly less training data. At the sequence level, \cite{Kim:2016aa} propose aligning the student's output distribution with the teacher's across entire sequences rather than individual tokens. We combine both ideas: sequence-level KL divergence with CoT supervision, weighted to stabilize label learning.

In a line of work related to ours, \cite{Cho:2019aa} compare distilled student performance under different architectures and find that larger or more accurate teachers do not necessarily produce better students. They conclude that student capacity is the limiting factor, more so than teacher size. Relatedly, \cite{Kaplun:2022aa} show that even poor-quality teachers can produce effective students, further suggesting that teacher size is not the binding constraint. However, these observations are made on models much smaller than those studied in our paper and only cover computer vision applications.

The topic of optimal compute allocation for distillation and the resulting scaling laws are analyzed in \cite{Busbridge:2025aa}, showing that distillation outperforms supervised pretraining up to a compute level that grows with student size. A statistical perspective on why distillation helps is presented in \cite{Menon:2020aa}, establishing a bias-variance tradeoff that explains how the teacher's class-probability estimates aid student learning.

Practical considerations for distillation at scale include online co-distillation for faster training via parallelism \cite{Anil:2018aa} and identification of key design choices for distillation quality \cite{Beyer:2021aa}, both at order-of-magnitude smaller models than ours. A complementary line of work studies on-policy KL distillation for language models, where the student samples its own outputs and the teacher scores them, addressing the train-inference exposure gap inherent in offline KL training~\cite{Agarwal:2024gkd}. More recently, distillation has become widely used in creating LLM families of different sizes, where the largest model is trained from scratch and acts as teacher to smaller variants \cite{Muralidharan:2024aa, Abdin:2024aa, Llama32:2024, DeepSeek-AI:2025aa, Bercovich:2025aa}. While distillation can be applied across different model families and tokenizers \cite{DeepSeek-AI:2025aa}, a common approach is to prune a larger model and then distill knowledge back into the pruned student. An extensive analysis of iterative pruning and knowledge distillation is presented in \cite{Muralidharan:2024aa}, which directly inspires our methodology.

From a complementary angle, \cite{Malach:2019aa} show that the benefit of depth and width at \emph{training} time is contingent on the target being approximable by shallower networks. We study how depth and width matter when \emph{compressing} already-trained models via pruning and distillation. Unlike \cite{Busbridge:2025aa}, who study scaling with compute allocation between teacher and student, we vary dataset size, compression ratio, and supervision format and report how in-domain and general-knowledge performance scale. Relative to \cite{Muralidharan:2024aa}, we add systematic comparison across data regimes and distillation methods (LoRA vs.\ logit-based, label vs.\ CoT).

Work on compression fragility shows that difficult downstream tasks are more sensitive to pruning and that damage from aggressive compression can be irreparable even with continued training \cite{Yin:2023aa}. We report analogous sensitivity but vary data size and distillation method rather than compression level and task difficulty.

Recent research demonstrates that LLMs can be effectively adapted to financial NLP tasks through careful fine-tuning, achieving strong results even with small models under 1.5B parameters \cite{rodriguez-inserte-etal-2023-large}. However, very small LLMs remain limited for tasks that benefit from richer reasoning or longer context \cite{Li:2025small}. Comparative studies such as \cite{Sharkey:2024aa} benchmark different architectures for financial sentiment analysis, finding that fine-tuned BERT models can outperform larger GPT models on domain-specific tasks, though encoder-only models have much shorter context windows and do not support generative or chain-of-thought use cases. Our work targets dense, decoder-only LLMs at multi-billion parameter scale with generative and chain-of-thought capabilities. Findings from encoder-only or sub-billion parameter models do not directly transfer to this setting.

Existing work on distillation scaling laws focuses on smaller models or single distillation methods, leaving open the question of how different distillation techniques compare at the scale of today's LLMs. Additionally, published research on knowledge distillation for financial applications remains sparse. \editgreen{This paper addresses both gaps. We compare logit-based and LoRA-based distillation with label-only and blended CoT supervision across data regimes at a fixed student size. We then study compression across student sizes using iterative blended CoT KD, with independently trained label-only LoRA models as non-iterative references.}

\section{Data}
\label{sec:data}

We aim for this work to serve as a guide for practitioners, and have therefore chosen to run experiments on data that reflects industry use cases rather than optimizing on academic benchmarks.

Due to the limited availability and diversity of open-source financial news datasets, we created a synthetic dataset of financial news. Headlines were generated with NVIDIA Nemotron-3-Nano-30B-A3B-BF16-Instruct~\cite{Nemotron3Nano30BA3B} using NVIDIA's NeMo Data Designer \cite{NeMoDataDesigner:2025}, then we applied fuzzy deduplication with NeMo Curator \cite{Jennings_NeMo-Curator_a_toolkit}. The deduplicated pool contains 500,000 headlines. Event labels are assigned by the Qwen3-32B teacher in \emph{thinking mode} (internal reasoning before the final class). The full corpus yields approximately 2.6 billion training tokens (counting full training sequences, i.e.\ prompt plus CoT response, rather than raw input text alone), comparable in scale to ablation-level distillation runs in prior work \cite{Muralidharan:2024aa, Bercovich:2024aa}. Training, validation, and testing splits are consistent across all experiments; only the dataset formats are adapted to each distillation method. The dataset will be released with this paper at \href{https://huggingface.co/datasets/nvidia/FinHeadlineMix}{\texttt{FinHeadlineMix}} to facilitate future research.

\begin{figure}[htbp]
  \centering
  \includegraphics[width=1\linewidth]{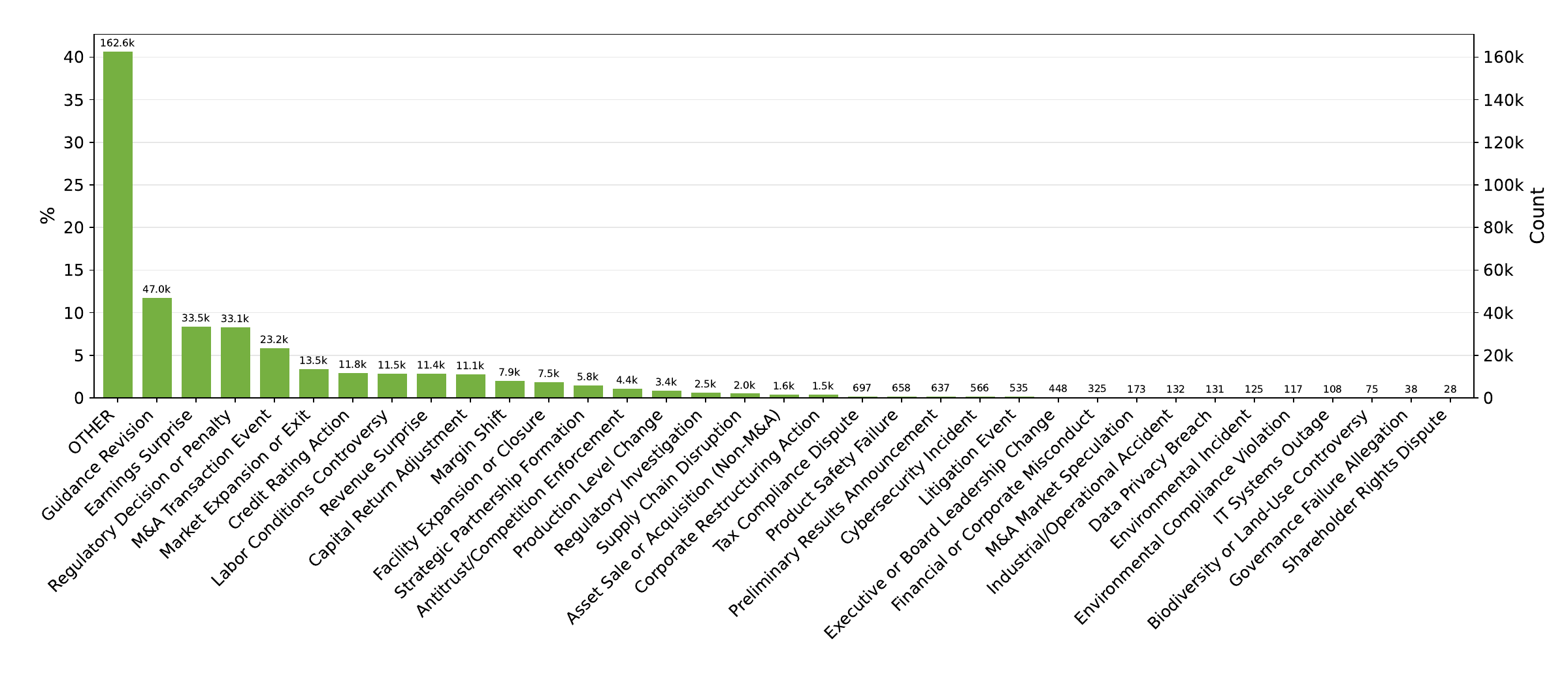}
  \caption{Train set label distribution over the 35 event classes in the financial news dataset (training split, classes sorted by decreasing frequency).}
  \label{fig:data_classes}
\end{figure}

The multiclass event identification task is a classification problem with 34 event types plus a ``other'' class, for 35 classes in total. Since headlines are generated independently of the class taxonomy, some may not fit neatly into predefined categories, which is a realistic scenario that mirrors real-world data. Headlines that match none of the defined classes are labeled as ``other.'' The dataset exhibits the class imbalance typical of such applications, as shown in Figure~\ref{fig:data_classes}. The class distribution is consistent across train, validation, and test splits. In practice, domain experts choose the task or taxonomy to match their target problem and can calibrate the prompt and teacher model on a small expert-labeled golden set (usually a few hundred examples; see Appendix~\ref{appendix:data}). Only this calibration set requires human annotation; the LLM handles the rest.

We selected Qwen3-32B \cite{Qwen3:2025} as the teacher, a dense 32B-parameter model with Grouped Query Attention (GQA)~\cite{ainslie2023gqa}, having found it comparable to Llama-3.1-405B \cite{llama3modelcard}, Llama-Nemotron-49B \cite{Bercovich:2025aa}, and GPT-OSS-120B \cite{OpenAI:GPToss:2025} on our in-domain event classification task. Its single-node footprint at inference simplifies deployment and compression compared to larger alternatives. We report results on this in-domain task and on general-knowledge benchmarks (Section~\ref{sec:experiments}) so practitioners can see the tradeoffs of distillation with domain-specific data versus performance on general benchmarks.

\textbf{Gold labels and evaluation.} Qwen3 exposes reasoning and non-reasoning inference in one architecture~\cite{Qwen3:2025}. With tensor parallelism ($\text{TP}=4$), greedy decoding without thinking produces short classification outputs (${\approx}\,$1--3 tokens) that are fully deterministic across runs; long \editgreen{native-thinking-mode} generations can differ across inference sessions because floating-point reductions in all-reduce operations across GPUs are not invariant to batch shape~\cite{He:2025nondeterminism}, and over hundreds to thousands of CoT tokens these differences accumulate and can shift the argmax. Naively comparing against a single set of gold labels therefore conflates model error with inference non-determinism: the teacher's \editgreen{native-thinking-mode} agreement with labels from a \emph{different} session is only ${\approx}\,$88\%, an artifact of TP non-determinism rather than a real accuracy ceiling. To eliminate this artifact, we generate evaluation gold labels from a single greedy non-thinking session, under which the teacher achieves 100\% accuracy (deterministic short outputs), providing a clean ceiling for student evaluation. Training data (CoT traces) are generated separately using the teacher's \editgreen{native thinking mode} in a single greedy session, with a small discard rate for failed generations (Appendix~\ref{appendix:teacher_eval}). \editgreen{All sections report results with Qwen's native thinking mode disabled; Section~\ref{sec:experiments} clarifies how learned CoT is handled.} Appendix~\ref{appendix:teacher_eval} details the relabeling setup and reproducibility.

\section{Methodology}
\label{sec:methods}

\subsection{Structured Pruning}

We apply joint depth and width pruning where the best configuration is chosen by lightweight NAS, as in \cite{Muralidharan:2024aa}. The lightweight NAS generates a set of candidate configurations that match the target model size, scores each using block importance and activation-based metrics for depth and width pruning respectively, and selects the configuration with the best quality. Combining both dimensions yields better compression-performance trade-offs than either approach alone. Other pruning configurations are theoretically possible, but more advanced NAS techniques \cite{Bercovich:2024aa} are outside our current scope. We restrict to pruned formats that remain hardware-optimal so that standard dense kernels can be used at inference without per-model custom CUDA code (see Appendix~\ref{appendix:prune}). After pruning, distillation is used to recover performance.

\textbf{Width pruning} selectively removes attention heads and neurons, maintaining the
layer-wise structure of the model. The lightweight NAS algorithm searches for optimal subnets by identifying and removing components that contribute least to performance using proxy metrics (attention entropy, weight norms, or induced loss). Let $w$ denote an architecture configuration and $L(w)$ the associated validation loss. The search solves
\begin{equation}
w^{\ast} = \underset{w \in \mathcal{W}}{\operatorname{argmin}} \left[\, L(w) + \lambda\, \operatorname{size}(w) \,\right]
\end{equation}
where $\mathcal{W}$ is the search space of candidate width configurations and $\lambda$ controls the trade-off between accuracy and model size.

\textbf{Depth pruning} removes entire Transformer blocks that contribute least to information propagation across hidden states. We quantify block significance using the Block Influence (BI) score from \cite{Men:2024aa}:

\begin{equation}
\mathrm{BI}_i = 1 - \mathbb{E}_{x \sim \mathcal{D},\, t} \left[
  \operatorname{cos}\left(X_{i, t},\, X_{i+1, t}\right)
\right]
\end{equation}
where
\begin{equation}
\operatorname{cos}\left(X_{i, t},\, X_{i+1, t}\right) =
\frac{ X_{i, t} \cdot X_{i+1, t} }{ \| X_{i, t} \|_2\, \| X_{i+1, t} \|_2 }
\end{equation}
denotes cosine similarity between hidden states before and after block $i$ at token position $t$. Blocks with low $\mathrm{BI}_i$ values induce minimal change in representations and are pruned first.

We use a balanced allocation between depth and width pruning. Concentrating width pruning on attention or key/value heads heavily degraded recoverable accuracy after distillation, and we avoid it (details in Appendix~\ref{appendix:what_didnt_work}). Appendix~\ref{appendix:prune} gives architectural details and practical guidance. Experimental results are in Section~\ref{sec:experiments}.

\subsection{Knowledge Distillation}

\editgreen{We compare four configurations along two axes: \emph{method} (LoRA vs.\ logit-based) and \emph{supervision format} (single-label vs.\ blended CoT).} Qwen3-32B \cite{Qwen3:2025} serves as the teacher throughout. Students are pruned versions at varying compression ratios.

\subsubsection*{LoRA distillation}
LoRA-based distillation~\cite{Hu:2021aa} trains the student to match the teacher's outputs (labels or CoT) using a standard cross-entropy loss, analogous to supervised fine-tuning (SFT). Rather than updating all model weights, LoRA freezes the pretrained parameters and injects trainable low-rank adapters into each layer.

Specifically, LoRA parameterizes an update to a weight matrix $W\in\mathbb{R}^{d_{\text{out}}\times d_{\text{in}}}$ as
\[
W_{\text{eff}} \;=\; W \;+\; \Delta W,\qquad \Delta W \;=\; \frac{\alpha_{\text{LoRA}}}{r}\, B A,\;\; B\in\mathbb{R}^{d_{\text{out}}\times r},\;\; A\in\mathbb{R}^{r\times d_{\text{in}}},
\]
where $r \ll \min(d_{\text{in}}, d_{\text{out}})$ is the adapter rank, controlling capacity, and $\alpha_{\text{LoRA}}$ is a scaling factor (we use $\alpha_{\text{LoRA}}=16$) that sets the magnitude of the adapter update. For label-only LoRA, we apply adapters to the query, key, and value projections in attention layers and to the gate and down projections in MLP layers, following recommendations from \cite{schulman2025lora}. In line with that work, we choose rank values adapted to dataset size so that the adapter is not capacity-limited (LoRA parameter count should remain above the effective information in the dataset, e.g.\ output tokens or samples times bits per label)~\cite{schulman2025lora}; rank scales from $r=4$ for the smallest datasets to $r=64$ for 400k samples. The blended-LoRA data-scaling runs use $r=32$ at every dataset size, target all linear projections, and optimize the same $\lambda=0.5$ label/CoT mean-loss mixture defined below.

\subsubsection*{Logits distillation}
Logits distillation uses the teacher's final-layer logits (the pre-softmax activations over the vocabulary) to train the student to match the teacher's probability distributions by minimizing Kullback--Leibler (KL) divergence~\cite{Kullback1951}. Unlike LoRA-based distillation, which only uses discrete outputs, logit-based distillation provides a richer signal by exposing the full distribution over the vocabulary.

Formally, let $p_{t,k}(x, \tau)$ and $p_{s,k}(x, \tau)$ denote the teacher and student output distributions at position $k$ for input $x$ with softmax temperature $\tau$. The distillation loss is:
\[
\mathcal{L}_{\mathrm{KL}} = \frac{1}{L} \sum_{k=1}^L \mathrm{KL}\Big( p_{t,k}(x, \tau)\,\|\,p_{s,k}(x, \tau) \Big),
\]
where $L$ is the sequence length and $k$ indexes token positions.

Prior work has explored alternative supervision targets such as ground-truth cross-entropy, reverse KL divergence, or intermediate hidden-state alignment~\cite{Gu:2023aa, Sun:2019aa}, as well as on-policy training where the student generates its own outputs for the teacher to score~\cite{Agarwal:2024gkd}. Following~\cite{Muralidharan:2024aa}, we focus exclusively on offline output-layer KL divergence minimization, which is both effective for accuracy recovery and simpler to implement; on-policy generation would also substantially increase per-step compute given the iterative pruning pipeline. We restrict supervision to final logits in all experiments unless otherwise noted.

\subsubsection*{Label-based supervision}

In label-based supervision, the teacher annotates each input with a single-class label and the student minimizes log-loss on these outputs (prompts and data formats in Appendix~\ref{appendix:data}).

\subsubsection*{CoT supervision}

In CoT~\cite{Wei:2022aa} supervision, the teacher generates a full CoT trace for each input, and the student is trained to reproduce the entire sequence, which can reduce the number of required training samples compared to label-only distillation~\cite{Hsieh:2023aa}.

Formally, let $(x, y_{1:L})$ denote an input $x$ paired with a CoT trace of $L$ tokens $y_1, \ldots, y_L$ generated by the teacher. The student is trained to minimize cross-entropy over the full trace, encouraging it to internalize the teacher's reasoning patterns in addition to answer accuracy.

For both label-based and CoT supervision, we mask the input tokens so that only the teacher's output serves as the training target.

\subsubsection*{Blended CoT}
\label{sec:blended_loss}
The teacher produces a CoT trace and a final label in a single response. When we train on the full response without explicitly weighting the loss on the label tokens, the gradient is spread over many diffuse CoT tokens and training is unstable, so the student does not reliably learn to classify. \editgreen{The per-token loss is cross-entropy for LoRA and KL divergence for logit-based distillation. We weight this loss as a convex combination of the mean loss on label tokens and the mean loss on CoT tokens.} We refer to this setup as \emph{blended CoT} in the rest of the paper. Formally:
\[
\mathcal{L} = \lambda\,\overline{\ell}_{\text{label}} + (1-\lambda)\,\overline{\ell}_{\text{cot}}.
\]
Thus $\lambda$ controls the share of the total loss from each region; $\lambda=0.5$ assigns half to the label part and half to the trace, regardless of how many tokens each region contains. We set per-token weights
\[
w_{\text{label}} = \lambda\, n_{\text{tot}}/n_{\text{l}}, \qquad w_{\text{cot}} = (1-\lambda)\, n_{\text{tot}}/n_{\text{c}}, \qquad n_{\text{tot}} = n_{\text{l}}+n_{\text{c}}.
\]
The factors $n_{\text{tot}}/n_{\text{l}}$ and $n_{\text{tot}}/n_{\text{c}}$ normalize by pool size so that each pool contributes its mean loss and $\lambda$ cleanly interpolates between the two means; the weight sum equals $n_{\text{tot}}$ for the framework's reduction. At the extremes, $\lambda=1$ uses only label positions and $\lambda=0$ only the trace. The forward pass always runs over the full response (CoT and label) for every $\lambda$; only the loss weights change. Because attention spans the full sequence, the teacher's KD targets at label positions condition on the preceding CoT, providing richer supervision than label-only or CoT-only training where the sequence or objective differs. To our knowledge, this is the first sequence-level KL-divergence formulation for decoder-only LLM distillation at multi-billion parameter scale that explicitly reweights label vs.\ trace regions: per-token reweighting is standard practice in LM cross-entropy training, but pairing it with KL-divergence supervision over CoT traces in this setting has not, to our knowledge, been previously reported. We sweep $\lambda \in \{0.01, 0.1, 0.5, 0.9, 0.99\}$, spanning near-trace-only ($\lambda{=}0.01$) to near-label-only ($\lambda{=}0.99$), and use $\lambda=0.5$, which we found most stable across experiments.

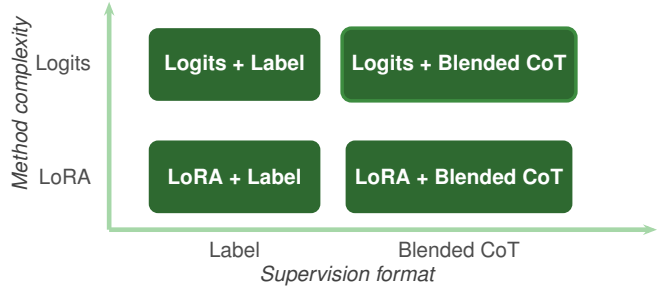
\begin{figure}[h]
  \centering
  \resizebox{0.62\textwidth}{!}{%
\begin{tikzpicture}[
  cell/.style={rectangle, rounded corners=4pt, draw=none,
    minimum width=2.6cm, minimum height=1.1cm,
    align=center, font=\sffamily\small\bfseries, inner sep=4pt},
  stable/.style  ={cell, fill=nvidiagreen!60!black, text=white},
  best/.style    ={cell, fill=nvidiagreen!60!black, text=white,
    draw=nvidiagreen!80!black, line width=1.5pt},
]
\def\cA{2.8} \def\cB{6.2}
\def\rA{1.25} \def\rB{2.95}
\node[stable]   at (\cA, \rA) {LoRA + Label};
\node[stable]   at (\cB, \rA) {LoRA + Blended CoT};
\node[stable]   at (\cA, \rB) {Logits + Label};
\node[best]     at (\cB, \rB) {Logits + Blended CoT};
\draw[-{Stealth[length=6pt]}, line width=2pt, color=nvidiagreen!50]
  (0.9, 0.45) -- (9.2, 0.45);
\draw[-{Stealth[length=6pt]}, line width=2pt, color=nvidiagreen!50]
  (0.9, 0.45) -- (0.9, 3.9);
\node[font=\sffamily\small\itshape, text=black!75]
  at (4.5, -0.25) {\editgreen{\textit{Supervision format}}};
\node[font=\sffamily\small\itshape, text=black!75, rotate=90, anchor=south]
  at (-0.15, 2.35) {\textit{Method complexity}};
\node[font=\sffamily\footnotesize, text=black!70] at (\cA, 0.15) {Label};
\node[font=\sffamily\footnotesize, text=black!70] at (\cB, 0.15) {Blended CoT};
\node[font=\sffamily\footnotesize, text=black!70, anchor=east] at (0.75, \rA) {LoRA};
\node[font=\sffamily\footnotesize, text=black!70, anchor=east] at (0.75, \rB) {Logits};
\end{tikzpicture}%
}
\caption{\editgreen{Four distillation configurations by method and supervision format.}}
\label{fig:distillation_grid}
\end{figure}

\editgreen{Blended CoT incorporates CoT effectively and yields better data efficiency (Section~\ref{sec:experiments}; in line with findings in~\cite{Hsieh:2023aa}). Training on CoT alone, without the blended weighting, was unstable in our experiments. The blended weighting yields stable one-epoch training for both LoRA and logit-based distillation; blended-LoRA inference additionally follows the learned-CoT protocol described at the start of Section~\ref{sec:experiments}.}

\subsection{Iterative Pruning and Distillation}
\label{sec:iterative}

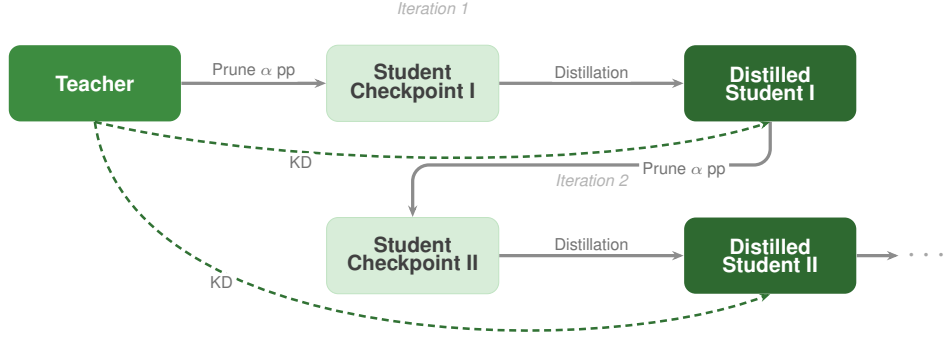
\begin{figure}[htbp]
\centering
\resizebox{0.9\textwidth}{!}{%
\begin{tikzpicture}[
  mbox/.style={
    rectangle, rounded corners=5pt, draw=none,
    minimum width=2.6cm, minimum height=1.15cm,
    align=center, font=\sffamily\small\bfseries, inner sep=4pt},
  teacher/.style ={mbox, fill=nvidiagreen!80!black, text=white},
  ckpt/.style    ={mbox, fill=nvidiagreen!22, text=black!75,
                   draw=nvidiagreen!35, line width=0.4pt},
  student/.style ={mbox, fill=nvidiagreen!60!black, text=white},
  flow/.style={-{Stealth[length=5pt,width=4pt]},
               line width=1.4pt, color=black!45},
  kd/.style  ={-{Stealth[length=4.5pt,width=3.5pt]},
               line width=1.1pt, color=nvidiagreen!65!black, densely dashed},
  op/.style={font=\sffamily\scriptsize, text=black!55,
             fill=white, inner sep=1.5pt, rounded corners=1pt},
]

\node[teacher] (T)  at (0,    0) {Teacher};
\node[ckpt]    (C1) at (4.8,  0) {Student\\[-2pt]Checkpoint\;I};
\node[student] (S1) at (10.2, 0) {Distilled\\[-2pt]Student\;I};

\node[ckpt]    (C2) at (4.8, -2.6) {Student\\[-2pt]Checkpoint\;II};
\node[student] (S2) at (10.2,-2.6) {Distilled\\[-2pt]Student\;II};
\node[font=\Large, text=black!35] (dots) at (12.6,-2.6) {$\cdots$};

\draw[flow] (T.east)  -- node[op, above] {Prune\;$\alpha$ pp} (C1.west);
\draw[flow] (C1.east) -- node[op, above] {Distillation}       (S1.west);

\draw[flow, rounded corners=7pt]
  (S1.south) -- ++(0,-0.65) -| (C2.north);
\node[op] at ($(S1.south)!0.5!(C2.north) + (1.4,0)$) {Prune\;$\alpha$ pp};

\draw[flow] (C2.east) -- node[op, above] {Distillation} (S2.west);
\draw[flow] (S2.east) -- (dots.west);

\draw[kd] (T.south) .. controls (3.0,-1.3) and (8.0,-1.3) ..
  node[op, below, pos=0.3] {KD} (S1.south);
\draw[kd] (T.south) .. controls (0.5,-4.0) and (7.5,-4.2) ..
  node[op, below, pos=0.3] {KD} (S2.south);

\node[font=\sffamily\scriptsize\itshape, text=black!30]
  at ($(T.north east)!0.5!(S1.north west) + (0,0.55)$) {Iteration 1};
\node[font=\sffamily\scriptsize\itshape, text=black!30]
  at ($(C2.north west)!0.5!(S2.north east) + (0,0.55)$) {Iteration 2};

\end{tikzpicture}%
}
\caption{Uniform iterative pruning and knowledge distillation: at each stage the model is pruned by $\alpha$ percentage points and then distilled from the original teacher.}
\label{fig:pruning-distillation-schema}
\end{figure}

We investigate an iterative pruning and distillation strategy for compressing LLMs. Figure~\ref{fig:pruning-distillation-schema} illustrates this process for \emph{logit-based} distillation. For LoRA we use a different procedure (per-size training), because the iterative approach is not applicable (see below). Starting from the teacher, we reduce model size by a constant step size in percentage points (pp; e.g.\ 21\,pp per step for the 4-step schedule in Table~\ref{tab:uniform_steps}) to produce intermediate student checkpoints. At each stage, the pruned model is refined via knowledge distillation from the original teacher on our domain-specific training data (Section~\ref{sec:data}), then pruned again. This alternating process continues until the target compression ratio is reached or until knowledge distillation can no longer recover performance.
For logit-based distillation, we follow the iterative approach of~\cite{Muralidharan:2024aa}. Section~\ref{sec:experiments} compares this against a single-step baseline (prune directly to the target size, distill once) to quantify the benefit of gradual compression.

For LoRA-based distillation, the iterative approach is not directly applicable. Structured pruning changes weight matrix dimensions ($W \to W'$ with different shape), making previously learned adapters $(A, B)$ dimensionally incompatible. Even with explicit remapping, pruning removes heads and neurons, so the learned low-rank factors no longer correspond to the same subspaces. To ensure a fair comparison, we train dedicated LoRA adapters for each student size independently.

Finally, for the iterative logit-based process we introduce a decayed compression schedule inspired by learning rate schedulers: instead of uniform pruning steps, we take larger steps early and smaller steps toward the end. This allows more aggressive initial compression while preserving fine-grained control as the model approaches the target size.

Throughout this work we use the \emph{original} teacher as the distillation source at every stage rather than first fine-tuning on domain data. A fine-tuned teacher could provide a stronger starting point, but verifying that fine-tuning has not caused forgetting is non-trivial, and practitioners who have limited domain data or want to try many compression variants cannot afford to re-fine-tune the teacher each time. The general-purpose teacher thus reflects the practical scenario we aim to study. Section~\ref{sec:experiments} quantifies this cost via a self-distillation baseline, showing that the dominant source of the student-to-teacher gap is distribution shift from training, not compression.

\paragraph{Implementation.} We use and adapt NeMo \cite{Harper_NeMo_a_toolkit} and NVIDIA Model Optimizer \cite{nvidia2025modelopt} implementations, with AdamW~\cite{Loshchilov:2019aa} and the default cosine annealing learning-rate schedule~\cite{Loshchilov:2017aa}. Optimizer hyperparameters were tuned within this setup. Adaptive, schedule-free alternatives may further improve stability and convergence (Section~\ref{sec:conclusion}).

\section{Experiments and Scaling Law Results}
\label{sec:experiments}

Students are distilled on in-domain data only (Section~\ref{sec:data}) and evaluated both on the in-domain task and on MMLU~\cite{hendrycks2021measuring} and MMLU-Pro~\cite{MMLU-Pro:2024}, so that every result quantifies the tradeoff between task performance and general knowledge. MMLU-Pro additionally exercises multi-step reasoning, so we use the pair as a joint proxy for general-knowledge and reasoning retention. We first compare distillation methods in terms of data efficiency at a fixed compression ratio (Section~\ref{sec:data_scaling}), then study uniform iterative pruning across compression ratios (Section~\ref{sec:iterative_scaling}), and finally propose decayed iterative schedules that push achievable compression further (Section~\ref{sec:scheduling}). We define the in-domain evaluation metrics below.

\editgreen{The data-scaling experiments compare all four configurations in Figure~\ref{fig:distillation_grid} at a fixed 50\% student size: label-only LoRA, blended-LoRA, direct-label KD, and blended CoT KD. They vary both training scope (low-rank adapters vs.\ full transformer-block updates) and supervision format (label-only vs.\ blended CoT). Because the two LoRA configurations use different rank and target-module settings (Section~\ref{sec:methods}), their contrast probes rather than isolates the effect of supervision format.}

\paragraph{Evaluation in class space.}
Let the evaluation set be $\{(x_i, y_i)\}_{i=1}^N$ with gold labels $y_i \in \{0,\dots,C-1\}$. For each input $x_i$, the model yields a class distribution $\mathbf{p}_i = (p_{i,0}, \dots, p_{i,C-1})$ over the $C$ task classes (from a class head or from decoded label probabilities), with $\sum_{c=0}^{C-1} p_{i,c} = 1$. We report three metrics in this space.

\textbf{Gold-label negative log-likelihood (NLL)} is the mean negative log probability assigned to the true class:
\[
\text{NLL} \;=\; -\frac{1}{N} \sum_{i=1}^{N} \log p_{i,y_i}.
\]
Lower is better.

\textbf{Multi-class Brier score} is the mean squared error between predicted probabilities and the one-hot gold label. With $\mathbf{e}_{y_i} \in \{0,1\}^C$ the one-hot vector for class $y_i$ (entry 1 at index $y_i$, 0 elsewhere),
\[
\text{Brier} \;=\; \frac{1}{N} \sum_{i=1}^{N} \left\| \mathbf{p}_i - \mathbf{e}_{y_i} \right\|_2^2
\;=\; \frac{1}{N} \sum_{i=1}^{N} \sum_{c=0}^{C-1} \bigl( p_{i,c} - \mathbb{1}[c = y_i] \bigr)^2,
\]
where $\mathbb{1}[c = y_i]$ denotes the indicator: 1 when $c = y_i$, 0 otherwise. Lower is better. Both are proper scoring rules: they reward honest probabilities and penalize overconfidence on the wrong class.

Training uses cross-entropy or KL divergence to teacher outputs. Evaluation uses \emph{gold} labels and the full class distribution, not only the argmax. NLL and Brier reflect \emph{discrimination} (mass on the correct class) and \emph{calibration} (match to the true outcome). We say gold-label NLL to avoid confusion with training loss (e.g.\ CE to teacher or CoT). Together they capture both discrimination and calibration: a model can be right often but miscalibrated (good classification metric, poor NLL/Brier) or well calibrated but wrong too often (good NLL/Brier, poor classification metric).

The third metric, \textbf{Macro F1}, is the unweighted mean of per-class F1: $F1_{\mathrm{macro}} = \frac{1}{C}\sum_{c=0}^{C-1} F1_c$, where $F1_c$ is the F1 score for class $c$ (precision and recall in the usual way). It measures \emph{classification} quality and is class-balanced but ignores the shape of $\mathbf{p}_i$; full definitions are in Appendix~\ref{appendix:alternative_metrics}. We report Macro F1, NLL, and Brier here. Weighted F1 and accuracy are in Appendix~\ref{appendix:alternative_metrics}.

\paragraph{Self-distillation baseline.}
To separate the cost of compression from the cost of the training process itself, we run each distillation method on the \emph{unpruned} teacher using the same pipeline and 200k training examples, matching the data budget at the final distillation step of all iterative experiments and LoRA training runs (the LoRA baseline also uses the same hyperparameters as the data-scaling experiments). The resulting models are not compressed; any gap from 100\% reflects \emph{distribution shift} introduced by the training process alone, not by pruning. This gap is not necessarily a quality degradation: fine-tuning on domain data may improve practical task performance, but because we evaluate against the original teacher's outputs (our gold labels), any shift in the student's output distribution registers as a drop from perfect agreement.

Self-KD (logit-based, blended $\lambda{=}0.5$, the ceiling for blended CoT~KD) achieves 0.8467 Macro~F1. LoRA (rank~32, the ceiling for LoRA students) achieves 0.7055. On general-knowledge benchmarks both methods retain ${\approx}\,$82 MMLU (vs.\ the teacher's 83.61) and 57--59 MMLU-Pro (vs.\ 65.54), confirming that domain fine-tuning on the unpruned model does not materially damage general knowledge.

The Self-KD ceiling of 0.8467 Macro~F1 is only 85\% of the way to perfect agreement with the teacher. The best iteratively compressed students (16\% of the teacher) reach ${\approx}\,$0.77, a further gap of ${\approx}\,$0.08. Distribution shift from training thus accounts for ${\approx}\,$0.15 of the total gap, roughly twice the compression cost. All scores throughout this paper are reported as raw values. The self-distillation baselines appear as dotted reference lines in the figures: the gap from any student score to the baseline isolates compression cost, while the gap from baseline to perfect agreement reflects distribution shift from the training process.

\paragraph{\editgreen{Inference protocol.}}
\editgreen{As described in Section~\ref{sec:data}, evaluation gold labels are generated in a single greedy session with Qwen's native thinking mode disabled, under which the untuned teacher achieves 100\% accuracy. Throughout this paper, \emph{non-thinking} refers only to disabling Qwen's native chat-template mechanism; it does not suppress CoT learned during fine-tuning. Qwen's native thinking mode is a model-specific chat-template feature and should not be conflated with CoT learned as ordinary assistant text. Blended-LoRA uses the project template and autoregressively generates its learned CoT followed by the final label, which we extract for scoring. The blended-LoRA results in Section~\ref{sec:data_scaling} use this protocol, which stabilized evaluation relative to enabling Qwen's native thinking path. All fine-tuned LoRA checkpoints use this native-thinking-off setting.}

\subsection{Data Scaling Laws}
\label{sec:data_scaling}

In this subsection the student size is fixed at 50\% of the teacher: the same student checkpoint is used for all data-scaling experiments, so that we isolate the effect of dataset size and distillation method without confounding from the compression ratio. We take 50\% as a practical deployment target (Section~\ref{sec:iterative_scaling} later confirms that performance at 50\% is well-preserved across methods); what we mean by 50\% balanced depth and width pruning is defined in Appendix~\ref{appendix:prune}.

\editgreen{At small data budgets, blended CoT KD is already the strongest classifier, showing that dense token-level supervision improves sample efficiency. Under the learned-CoT inference protocol, blended-LoRA is stable across seeds and improves quickly as data are added before reaching a plateau. Its NLL and Brier follow a different scaling regime: they initially improve but then worsen as the CoT corpus grows, so additional data do not uniformly improve class-probability quality. Direct-label KD remains the least reliable configuration, with delayed and non-monotonic recovery.}

\editgreen{These trends show that dataset size cannot be interpreted independently of supervision format and update scope. Blended CoT KD converts additional examples into classification gains most consistently, whereas blended-LoRA increasingly trades class-probability quality for broader capability retention. All experiments in this subsection use three independent seeds.}

\begin{figure}[htbp]
  \centering
  \begin{minipage}[t]{0.33\linewidth}
    \centering
    \includegraphics[width=\linewidth]{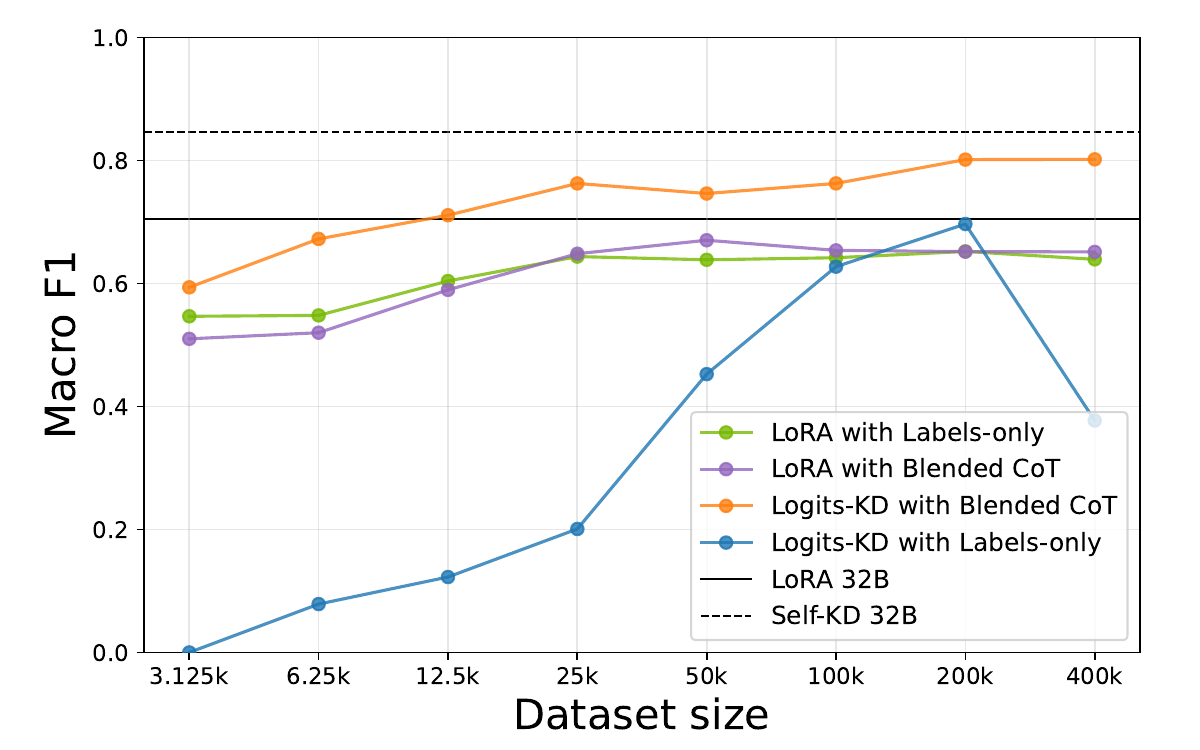}
    \smallskip
    \small (a) Macro F1.
  \end{minipage}\hfill
  \begin{minipage}[t]{0.33\linewidth}
    \centering
    \includegraphics[width=\linewidth]{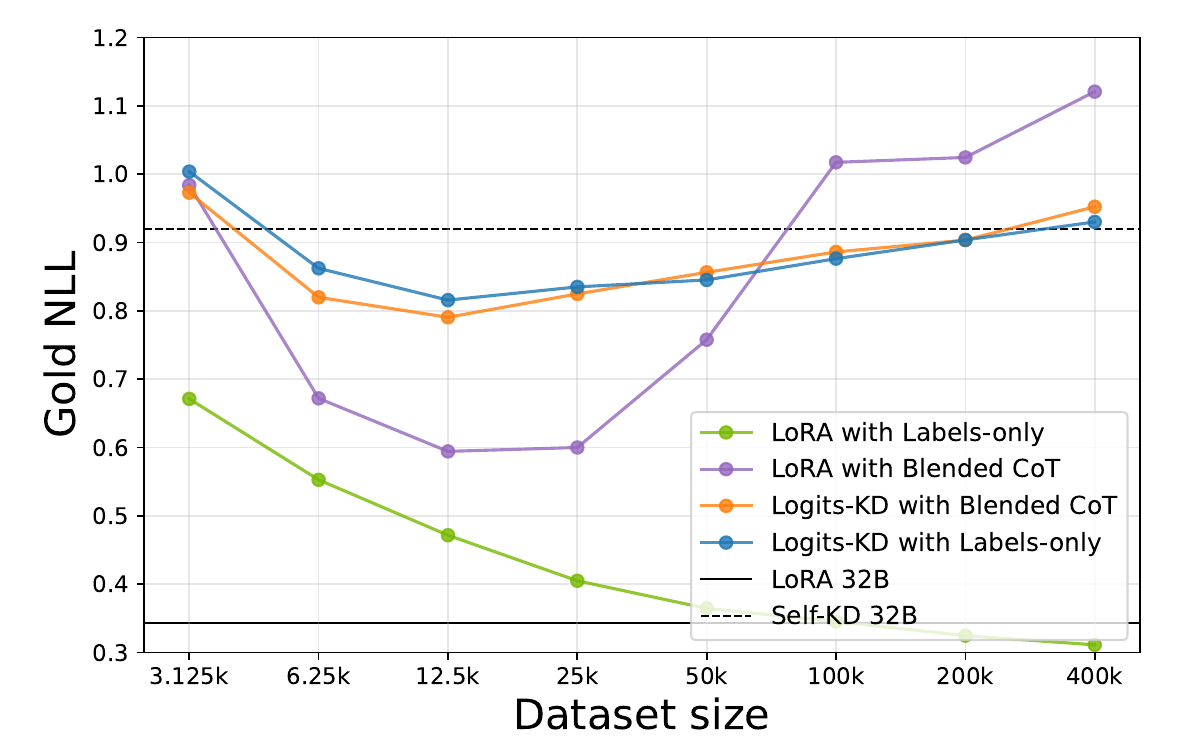}
    \smallskip
    \small (b) Gold-label NLL.
  \end{minipage}\hfill
  \begin{minipage}[t]{0.33\linewidth}
    \centering
    \includegraphics[width=\linewidth]{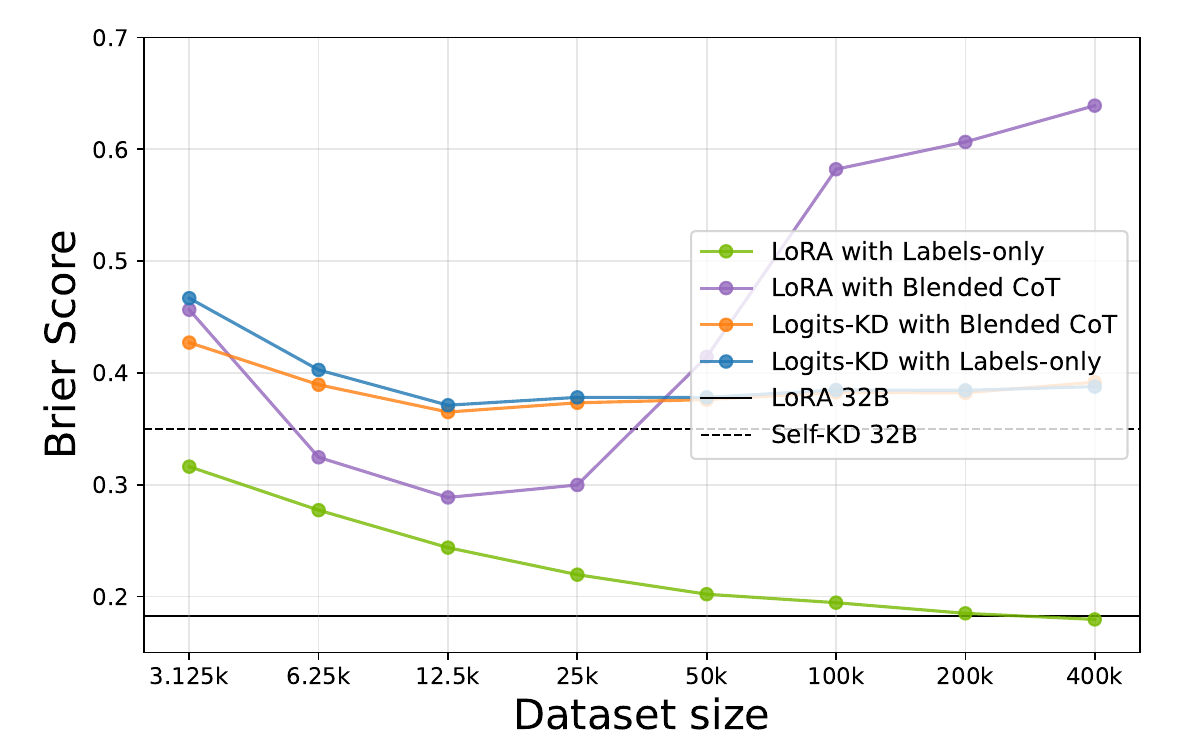}
    \smallskip
    \small (c) Brier score.
  \end{minipage}
  \caption{In-domain performance as a function of training dataset size for label-only LoRA, blended-LoRA, direct-label KD, and blended CoT KD. (a) Macro F1; (b) gold-label NLL; (c) Brier score. Mean across three seeds. Horizontal lines mark the Self-KD and LoRA baselines on the unpruned 32B model. Lower is better for NLL and Brier.}
  \label{fig:data_scaling_indomain}
\end{figure}

\editgreen{Blended CoT KD maintains the highest Macro F1 and continues to improve across the data range. Both LoRA variants improve rapidly and then plateau, suggesting that adding examples eventually becomes less important than the constraints of the low-rank update. Their NLL and Brier diverge: label-only LoRA improves with more data, whereas blended-LoRA deteriorates after its early optimum. The small cross-seed spread indicates a systematic tradeoff rather than unstable convergence. One possible explanation is that long CoT targets increasingly compete for limited adapter capacity; isolating this effect is left for future work. Direct-label KD remains fragile and non-monotonic. All LoRA results use a single epoch (Appendix~\ref{appendix:what_didnt_work}).}

\begin{figure}[htbp]
  \centering
  \begin{minipage}[t]{0.33\linewidth}
    \centering
    \includegraphics[width=\linewidth]{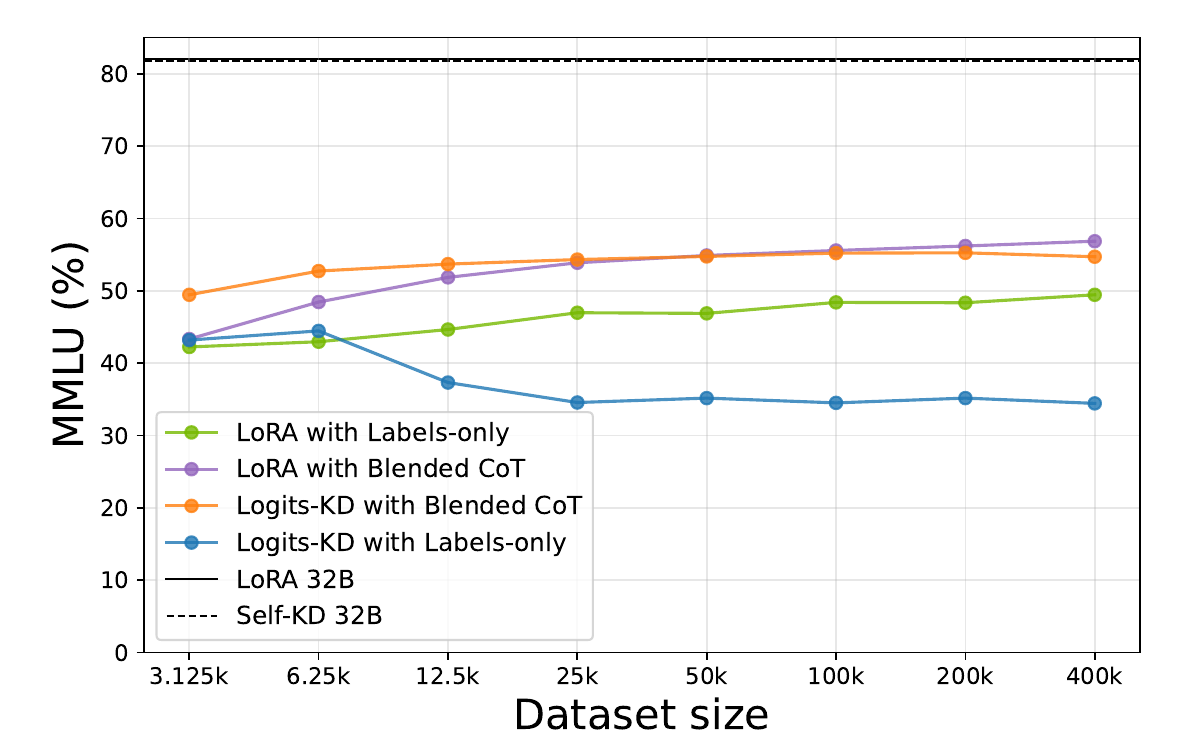}
    \smallskip
    \small (a) MMLU.
  \end{minipage}\hspace{2em}
  \begin{minipage}[t]{0.33\linewidth}
    \centering
    \includegraphics[width=\linewidth]{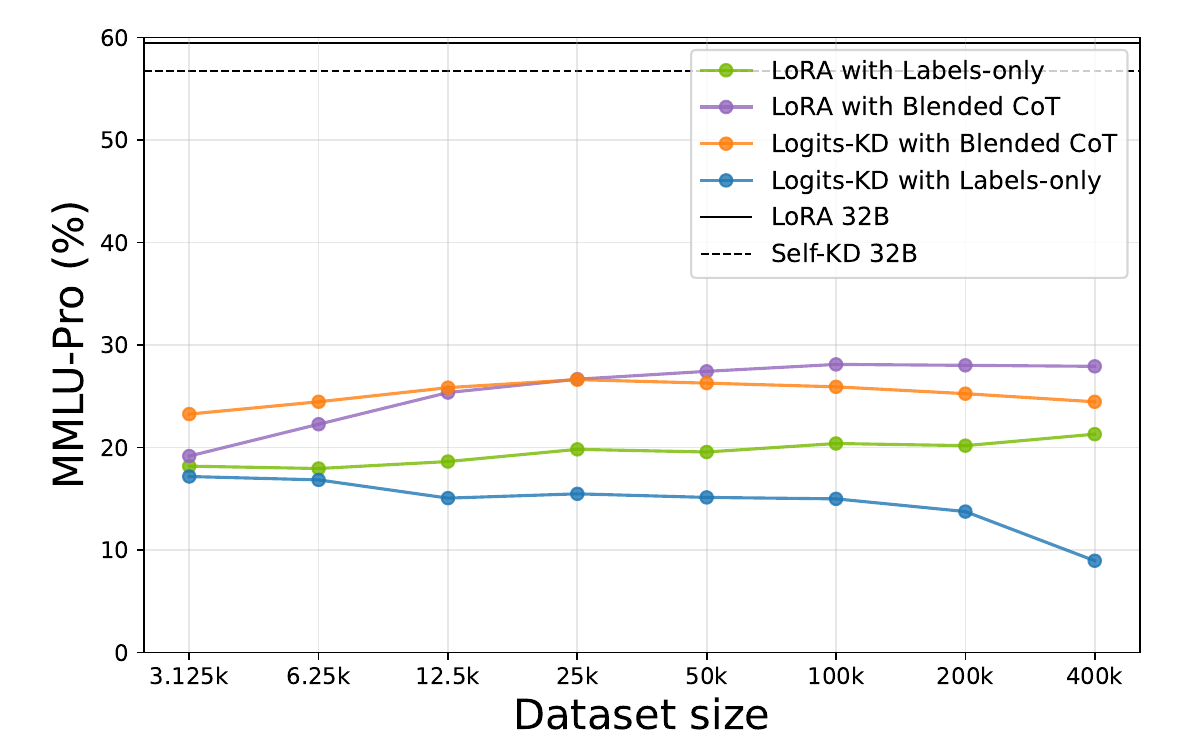}
    \smallskip
    \small (b) MMLU-Pro.
  \end{minipage}
  \caption{General-knowledge benchmarks for the four data-scaling configurations: MMLU and MMLU-Pro scores as a function of training dataset size for the 50\% student. Qwen3-32B-Base reference: 83.61 on MMLU and 65.54 on MMLU-Pro. The 50\% pruned model before distillation scores 40.30 on MMLU and 17.17 on MMLU-Pro.}
  \label{fig:data_scaling_mmlu}
\end{figure}

\paragraph{General-knowledge retention.}
We also evaluate the same models on general-knowledge benchmarks (MMLU, MMLU-Pro) as a comparison to the in-domain task (Figure~\ref{fig:data_scaling_mmlu}). The full teacher (Qwen3-32B-Base) reaches 83.61 on MMLU and 65.54 on MMLU-Pro~\cite{Qwen3:2025}. As a reference, the 50\% pruned model \emph{before any distillation} scores 40.30 on MMLU and 17.17 on MMLU-Pro; this baseline isolates the damage from pruning alone and allows us to separate the effect of each distillation method on general knowledge.

\editgreen{Label-only LoRA remains close to the pruned baseline, largely preserving rather than recovering the general knowledge that survives pruning. Blended CoT KD actively recovers general knowledge, suggesting that the reasoning trace anchors the teacher's broader distribution. Blended-LoRA shows the same benefit under adapter-based training and eventually surpasses blended CoT KD at the largest data budgets. Direct-label KD exhibits the opposite pattern, losing more general knowledge as in-domain data increase.}

\paragraph{Training scope vs.\ supervision format.}
\editgreen{The LoRA comparison is directional rather than fully controlled because the label-only and blended configurations differ in rank and target modules.}

\paragraph{Practitioner guidance.} Blended CoT KD is the strongest choice for in-domain classification and is the most sample-efficient. Label-only LoRA provides \editgreen{the lowest NLL and Brier}, especially at large data. Blended-LoRA is useful when adapter-only training and general-knowledge retention are priorities, but its NLL and Brier indicate that the data budget should be tuned rather than maximized. Direct-label KD should be avoided when general-knowledge preservation or stable convergence matters.

For in-domain metrics the teacher's outputs define the training targets, so the student approximates a ceiling it cannot exceed; for MMLU/MMLU-Pro the teacher's reference scores provide an upper bound. The 50\% student size therefore caps how much performance can recover; in Section~\ref{sec:iterative_scaling} we vary the compression ratio to study how it scales with model size.

\subsection{Uniform iterative compression scaling laws}
\label{sec:iterative_scaling}

\editgreen{Following Section~\ref{sec:data_scaling}, we drop direct-label KD and focus the compression-ratio study on blended CoT KD. Label-only LoRA is retained as a non-iterative reference: adapters are trained independently at each student size rather than propagated between pruning steps. Blended-LoRA is evaluated only in the fixed-size data-scaling study. Section~\ref{sec:scheduling} considers only the iterative logit-based pipeline.} Compression follows Table~\ref{tab:uniform_steps}: model size after each step when going from 100\% to 16\% in 1 to 4 uniform steps, with constant per-step reduction. LoRA is trained on 200k examples at every student size, evaluated at the same set of model sizes as the iterative blended CoT runs, but each LoRA student is trained independently from that pruned size.

\begin{figure}[htbp]
  \centering
  \begin{minipage}[t]{0.33\linewidth}
    \centering
    \includegraphics[width=\linewidth]{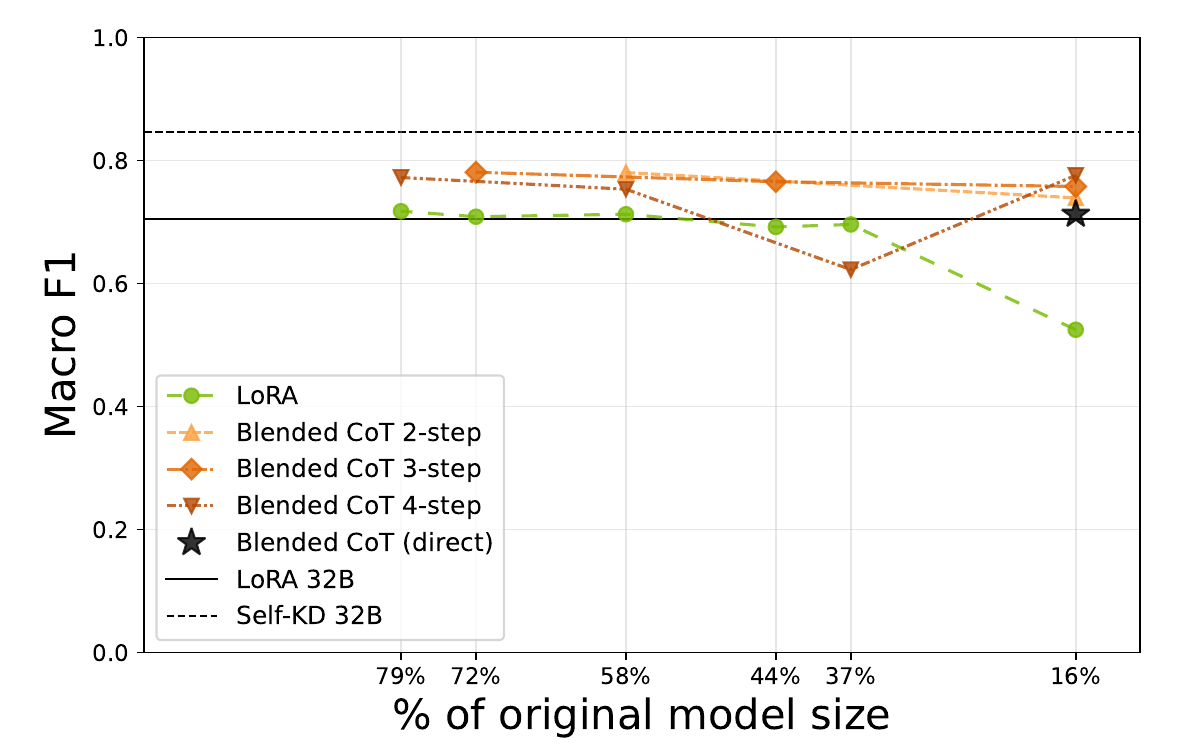}
    \smallskip
    \small (a) Macro F1.
  \end{minipage}\hfill
  \begin{minipage}[t]{0.33\linewidth}
    \centering
    \includegraphics[width=\linewidth]{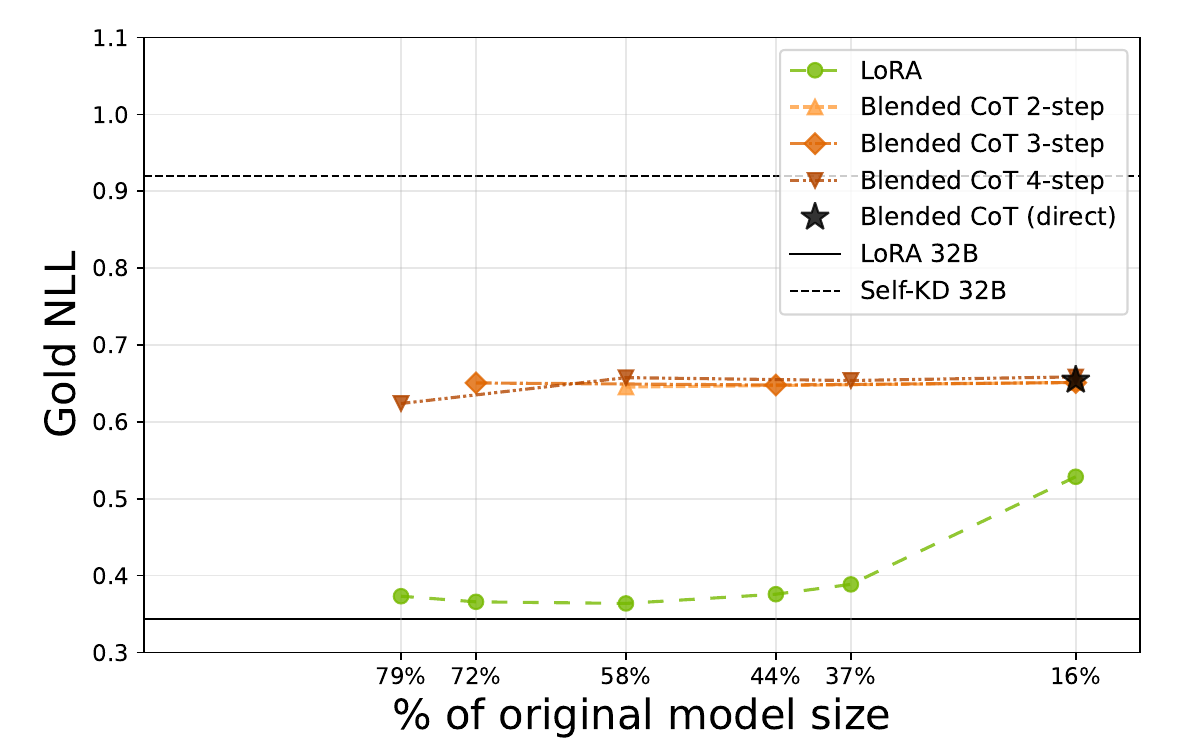}
    \smallskip
    \small (b) Gold-label NLL.
  \end{minipage}\hfill
  \begin{minipage}[t]{0.33\linewidth}
    \centering
    \includegraphics[width=\linewidth]{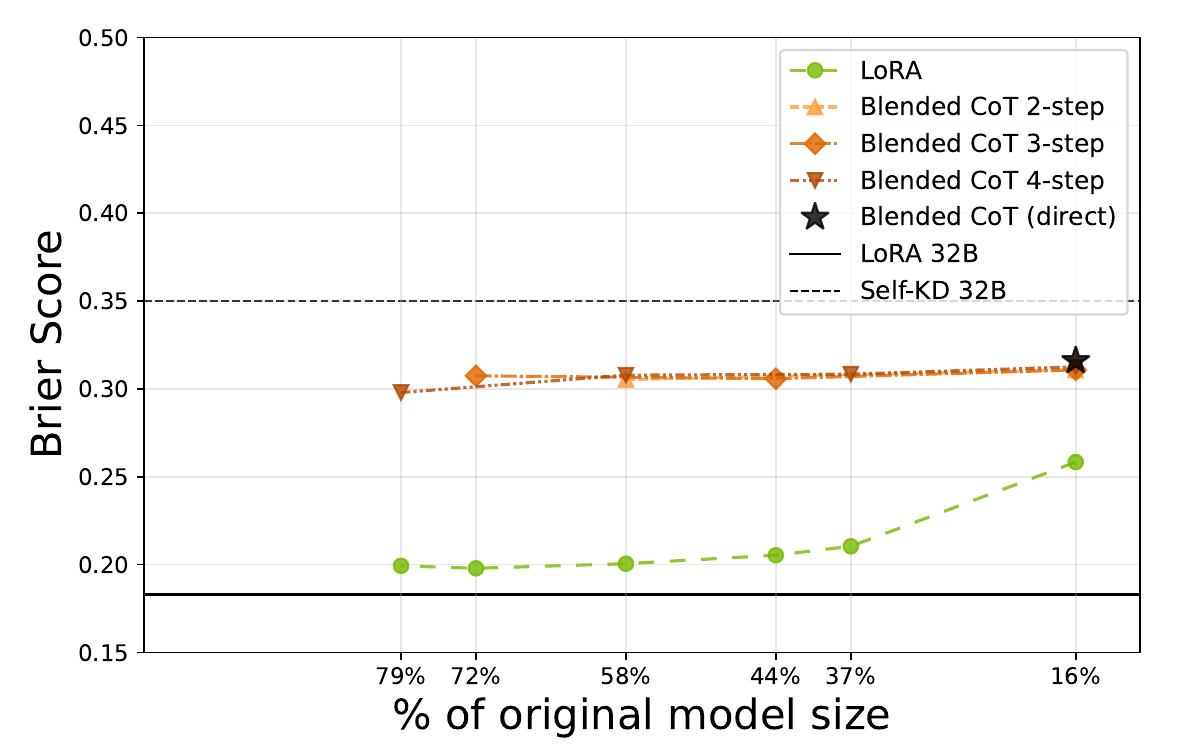}
    \smallskip
    \small (c) Brier score.
  \end{minipage}
  \caption{In-domain performance under uniform iterative pruning and distillation. (a) Macro F1; (b) gold-label NLL; (c) Brier score. Each panel shows the mean across three seeds; lower is better for NLL and Brier. Dotted horizontal lines mark the self-distillation baselines (Self-KD 200k and LoRA 200k, unpruned teacher): the gap above is distribution shift from training, below it is compression cost. LoRA uses 200k examples at each size; blended CoT KD uses a graduated data regime (see text). Compression schedule is given in Table~\ref{tab:uniform_steps}.}
  \label{fig:uniform_iter_indomain}
\end{figure}

For blended CoT KD, we use a graduated data regime that doubles the training set at each successive step, fixing the final step at 200k so that all schedules receive identical supervision at the 16\% target (e.g.\ 25k, 50k, 100k, 200k for the 4-step schedule). Earlier steps converge with less data because the model is still large; later steps need more data to recover from progressively aggressive pruning. We run all configurations with three seeds and report the mean in-domain score (Figure~\ref{fig:uniform_iter_indomain}).

\begin{table}[htbp]
\centering
\caption{Uniform pruning to 16\%: model size (\%) and per-step reduction (pp) when using 1, 2, 3, or 4 steps. Per-step reduction is 84\,pp divided equally; the 4-step schedule uses 21\,pp per step.}
\label{tab:uniform_steps}
\small
\begin{tabular}{@{}l*{4}{C{2.8em}}@{}}
\toprule
\textbf{Steps} & \textbf{Step 1} & \textbf{Step 2} & \textbf{Step 3} & \textbf{Step 4} \\
\midrule
\multicolumn{5}{@{}l}{\textit{Model size after step as fraction of teacher model (\%)}} \\
\midrule
1 step  & 16 & -- & -- & -- \\
2 steps & 58 & 16 & -- & -- \\
3 steps & 72 & 44 & 16 & -- \\
4 steps & 79 & 58 & 37 & 16 \\
\midrule
\multicolumn{5}{@{}l}{\textit{Reduction at step (pp)}} \\
\midrule
1 step  & 84 & -- & -- & -- \\
2 steps & 42 & 42 & -- & -- \\
3 steps & 28 & 28 & 28 & -- \\
4 steps & 21 & 21 & 21 & 21 \\
\bottomrule
\end{tabular}
\end{table}
\vspace{-0.5em}

\editgreen{Single-step pruning to small model sizes degrades recovery: at 16\% of the teacher, label-only LoRA reaches 0.5248 mean Macro F1 (down from 0.6960 at 37\%), while direct-to-16\% blended CoT KD reaches 0.7126. Iterative multi-step compression improves the blended CoT KD endpoint to 0.7392 with two steps, 0.7578 with three steps, and 0.7759 with four steps. The 4-step schedule therefore has the highest final mean, although its 37\% intermediate checkpoint is unstable across seeds. At moderate compression, blended CoT KD generally has higher Macro F1, whereas label-only LoRA has lower NLL and Brier; no method combines the highest Macro F1 with the lowest NLL/Brier.}

\begin{figure}[htbp]
  \centering
  \begin{minipage}[t]{0.33\linewidth}
    \centering
    \includegraphics[width=\linewidth]{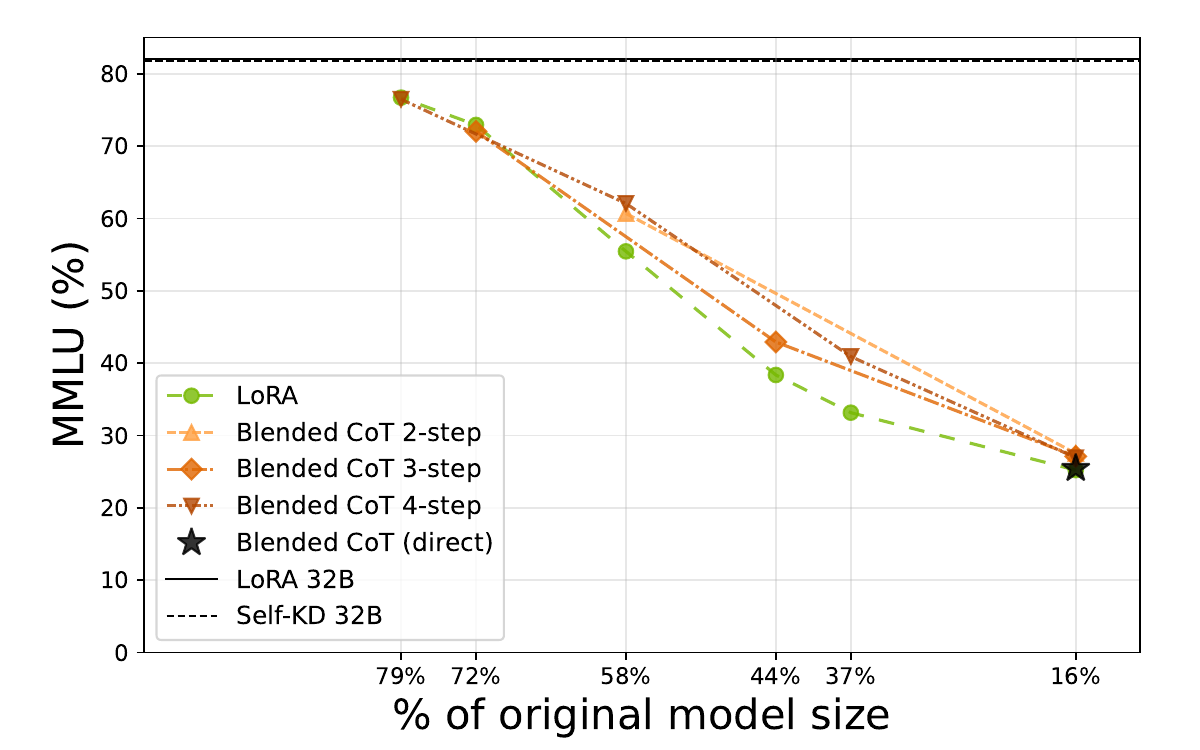}
    \smallskip
    \small (a) MMLU.
  \end{minipage}\hspace{2em}
  \begin{minipage}[t]{0.33\linewidth}
    \centering
    \includegraphics[width=\linewidth]{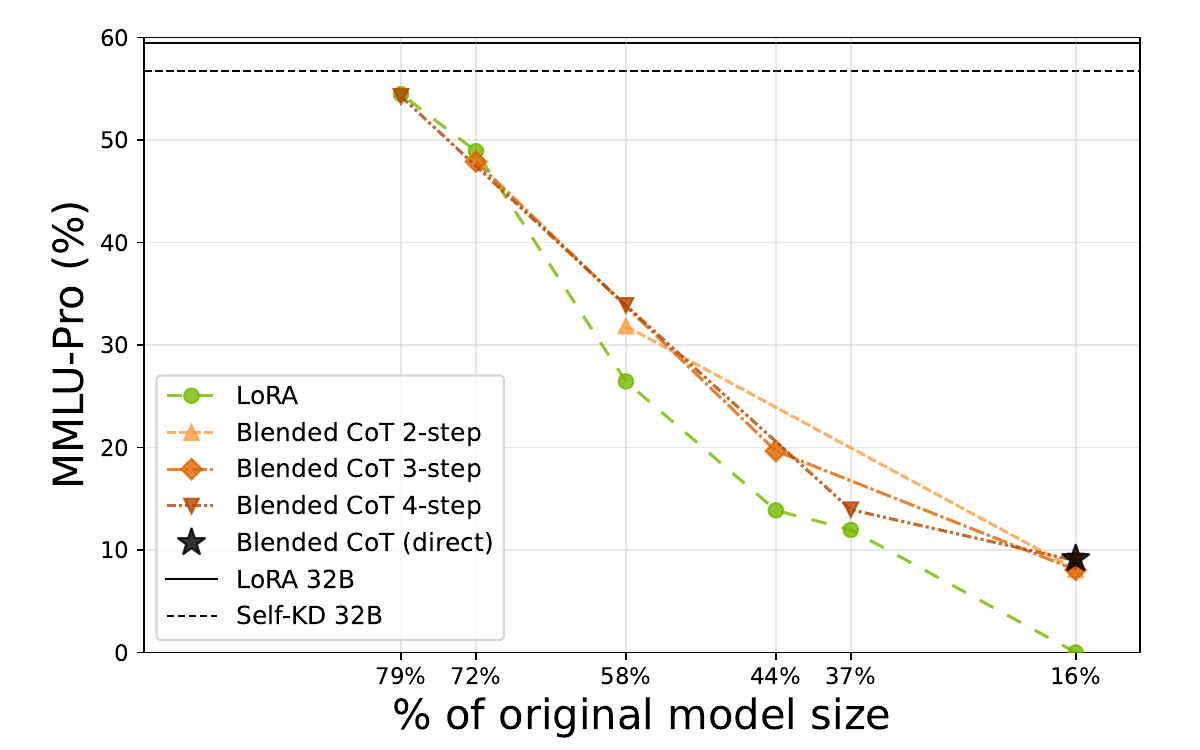}
    \smallskip
    \small (b) MMLU-Pro.
  \end{minipage}
  \caption{General-knowledge benchmarks for uniform iterative compression: MMLU and MMLU-Pro scores as a function of model size (compression ratio). LoRA is shown as a non-iterative reference (adapters independently trained at each size). Same data regimes and seeds (three per configuration) as in Figure~\ref{fig:uniform_iter_indomain}.}
  \label{fig:iterative_mmlu_mmlupro}
\end{figure}

We also evaluate the uniform iterative checkpoints on MMLU and MMLU-Pro (Figure~\ref{fig:iterative_mmlu_mmlupro}). The number of iterative steps has little effect on MMLU at the final target, suggesting that general-knowledge loss is governed primarily by the final model size rather than the compression path. Blended CoT KD retains substantially more general knowledge than LoRA at intermediate compression levels, consistent with the supervision-format effect observed in Section~\ref{sec:data_scaling}; the gap is negligible at large sizes but widens steadily with compression. At the 16\% target all methods converge near the random-guess baseline (${\approx}\,$25 on MMLU, ${\approx}\,$10--12 on MMLU-Pro), indicating that general knowledge is effectively lost at this compression level.

Two additional factors govern recovery quality beyond the number of steps. First, the pruning configuration at each target size matters: for a given compression ratio, different depth--width allocations are possible, and we find that balanced allocation is important for recoverability (Appendix~\ref{appendix:what_didnt_work}). Second, the per-step reduction must remain within a recoverable range: the 4-step schedule's higher variance at intermediate checkpoints (Figure~\ref{fig:uniform_iter_indomain}) suggests that even moderate single-step reductions of ${\approx}\,$21\,pp can occasionally exceed the recovery threshold. A natural question is whether front-loading the pruning, removing more when the model is large and redundant and less when it is capacity-constrained, can improve on uniform step sizes. Section~\ref{sec:scheduling} tests whether front-loaded, decayed schedules can improve on uniform steps at the same 16\% target and push compression further.

\subsection{Iterative compression scheduling}
\label{sec:scheduling}

We ask two questions: do decayed schedules preserve task performance better than uniform at the same final target, and do they allow pushing compression further? We use only logit-based KD with blended CoT (no LoRA), with the same graduated data regime as Section~\ref{sec:iterative_scaling} (doubling at each step, with 400k for the fifth step where applicable), three seeds, and mean reporting. The uniform row in all tables and figures reuses the 4-step blended CoT data from Section~\ref{sec:iterative_scaling}. We evaluate four decayed schedules:\footnote{Each schedule is fully determined by the target ratio $r_{\text{target}}$ and number of steps $N$; concrete sizes for all schedules are given in Table~\ref{tab:schedule_comparison}. Exact parametric forms and implementation will be provided in the accompanying code release.}
\begin{itemize}[leftmargin=2em, labelsep=0.45em, itemsep=2pt, parsep=0pt, topsep=4pt]
    \item \textbf{Exponential decay}: each step removes exponentially less than the previous.
    \item \textbf{Polynomial decay}: target sizes follow $(1 - t/T)^p$ with $p = 2$.
    \item \textbf{Cosine annealing}~\cite{Loshchilov:2017aa}: smooth cosine interpolation between start and target sizes.
    \item \textbf{Linear decay}: per-step reduction (in pp) decreases linearly.
\end{itemize}

To illustrate the differences, consider compressing a model from 100\% to 16\% of its original size over 4 iterative steps. Table~\ref{tab:schedule_comparison} shows how each schedule distributes the pruning across steps, and includes an extrapolated fifth step showing how much further each schedule could push compression.

\begin{table}[!htbp]
\centering
\caption{Comparison of pruning schedules: model size (\%) and per-step reduction (pp) for 4-step compression from 100\% to 16\%, with an extrapolated Step~5. All schedules reach 16\% at Step~4; Step~5 shows how much further each can push.}
\label{tab:schedule_comparison}
\small
\begin{tabular}{@{}l*{5}{C{2.8em}}@{}}
\toprule
\textbf{Schedule} & \textbf{Step 1} & \textbf{Step 2} & \textbf{Step 3} & \textbf{Step 4} & \textbf{Step 5} \\
\midrule
\multicolumn{6}{@{}l}{\textit{Model size after step as fraction of teacher model (\%)}} \\
\midrule
Uniform     & 79 & 58 & 37 & 16 & N/A \\
Linear      & 73 & 50 & 31 & 16 & 5  \\
Cosine      & 70 & 46 & 28 & 16 & 10 \\
Polynomial  & 61 & 36 & 22 & 16 & 14 \\
Exponential & 48 & 27 & 19 & 16 & 15 \\
\midrule
\multicolumn{6}{@{}l}{\textit{Reduction at step (pp)}} \\
\midrule
Uniform     & 21 & 21 & 21 & 21 & 21 \\
Linear      & 27 & 23 & 19 & 15 & 11 \\
Cosine      & 30 & 24 & 18 & 12 &  6 \\
Polynomial  & 39 & 25 & 14 &  6 &  2 \\
Exponential & 52 & 21 &  8 &  3 &  1 \\
\bottomrule
\end{tabular}
\end{table}
\vspace{-0.5em}

\begin{figure}[!htbp]
  \centering
  \begin{minipage}[t]{0.33\linewidth}
    \centering
    \includegraphics[width=\linewidth]{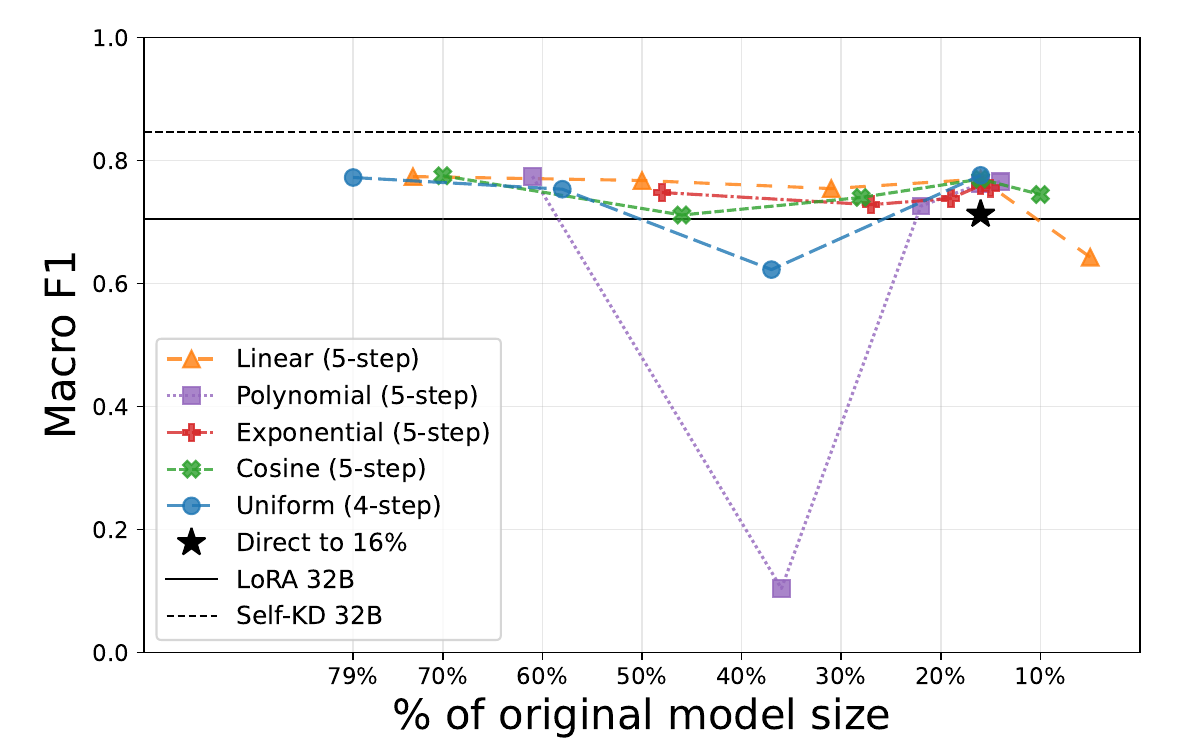}
    \smallskip
    \small (a) Macro F1.
  \end{minipage}\hfill
  \begin{minipage}[t]{0.33\linewidth}
    \centering
    \includegraphics[width=\linewidth]{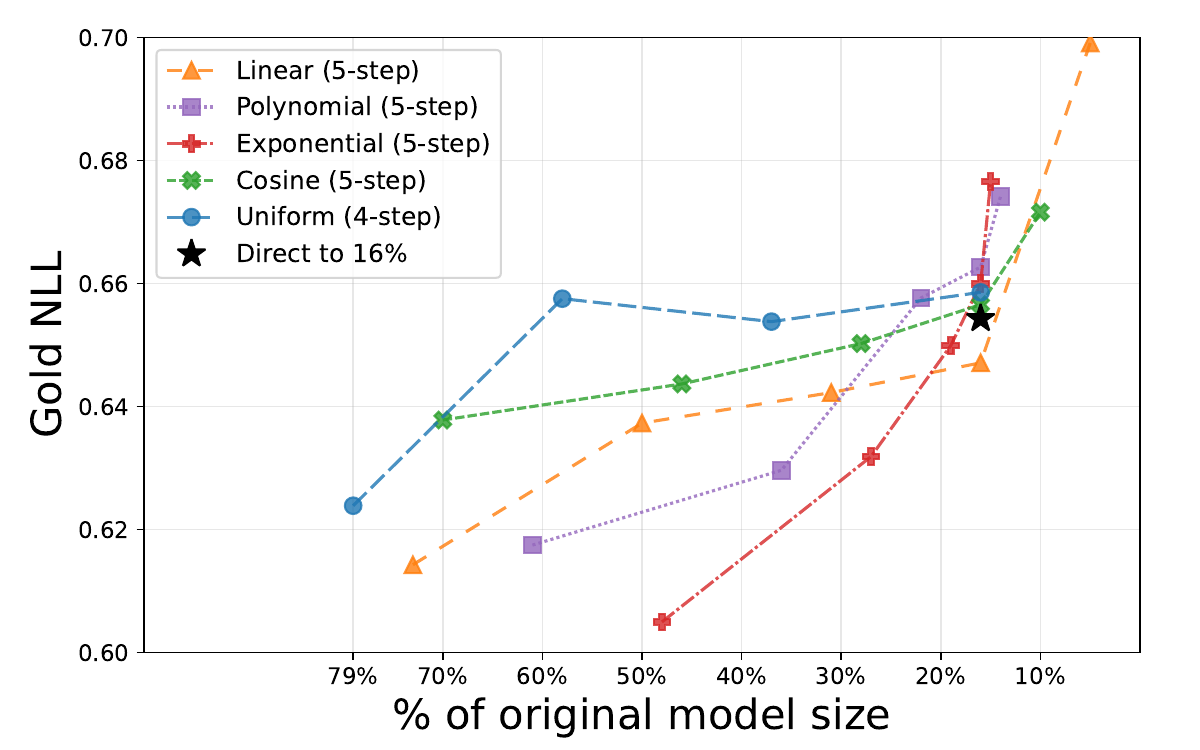}
    \smallskip
    \small (b) Gold-label NLL.
  \end{minipage}\hfill
  \begin{minipage}[t]{0.33\linewidth}
    \centering
    \includegraphics[width=\linewidth]{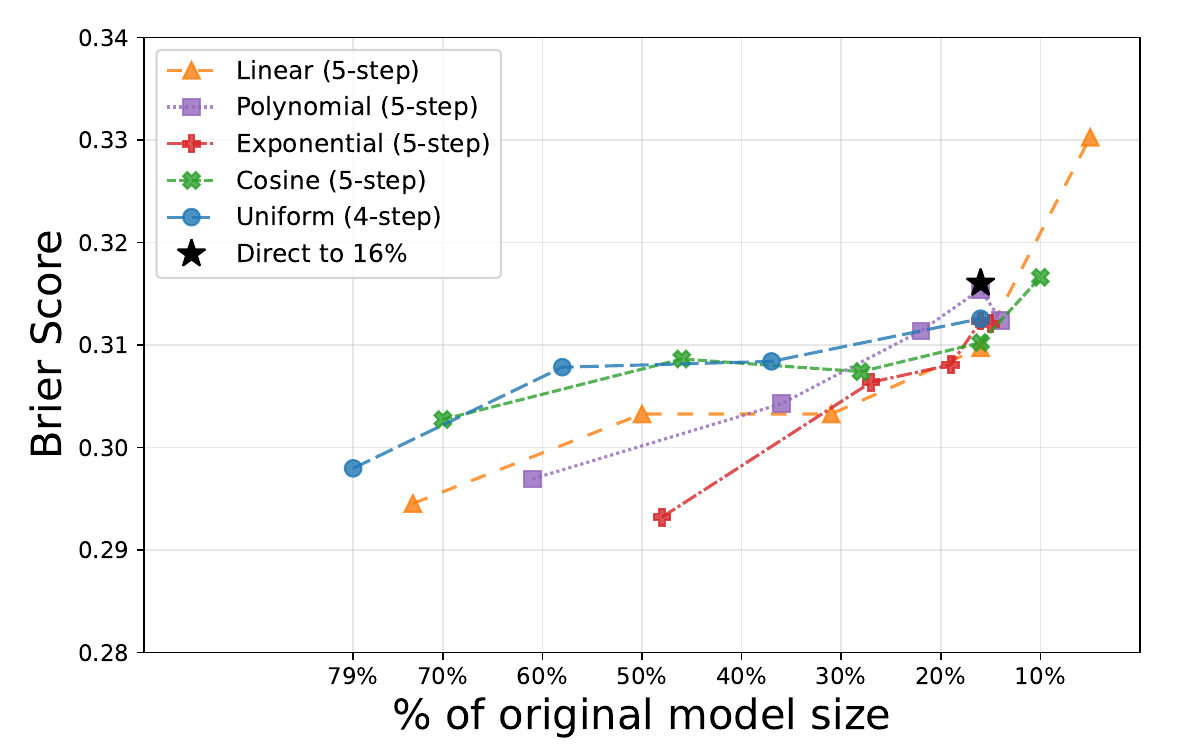}
    \smallskip
    \small (c) Brier score.
  \end{minipage}
  \caption{In-domain performance under decayed and uniform iterative schedules. (a) Macro F1; (b) gold-label NLL; (c) Brier score. Each panel shows the mean across three seeds at intermediate and final steps (Table~\ref{tab:schedule_comparison}) using the graduated data regime described in Section~\ref{sec:iterative_scaling}; lower is better for NLL and Brier. The gray star marks the direct-to-16\% single-step baseline (prune once, distill once). The dotted horizontal line marks the Self-KD 200k baseline (unpruned teacher, same pipeline): the gap from perfect agreement to this line is distribution shift, the gap below it is compression cost.}
  \label{fig:schedule_in_domain}
\end{figure}

\begin{figure}[!htbp]
  \centering
  \begin{minipage}[t]{0.33\linewidth}
    \centering
    \includegraphics[width=\linewidth]{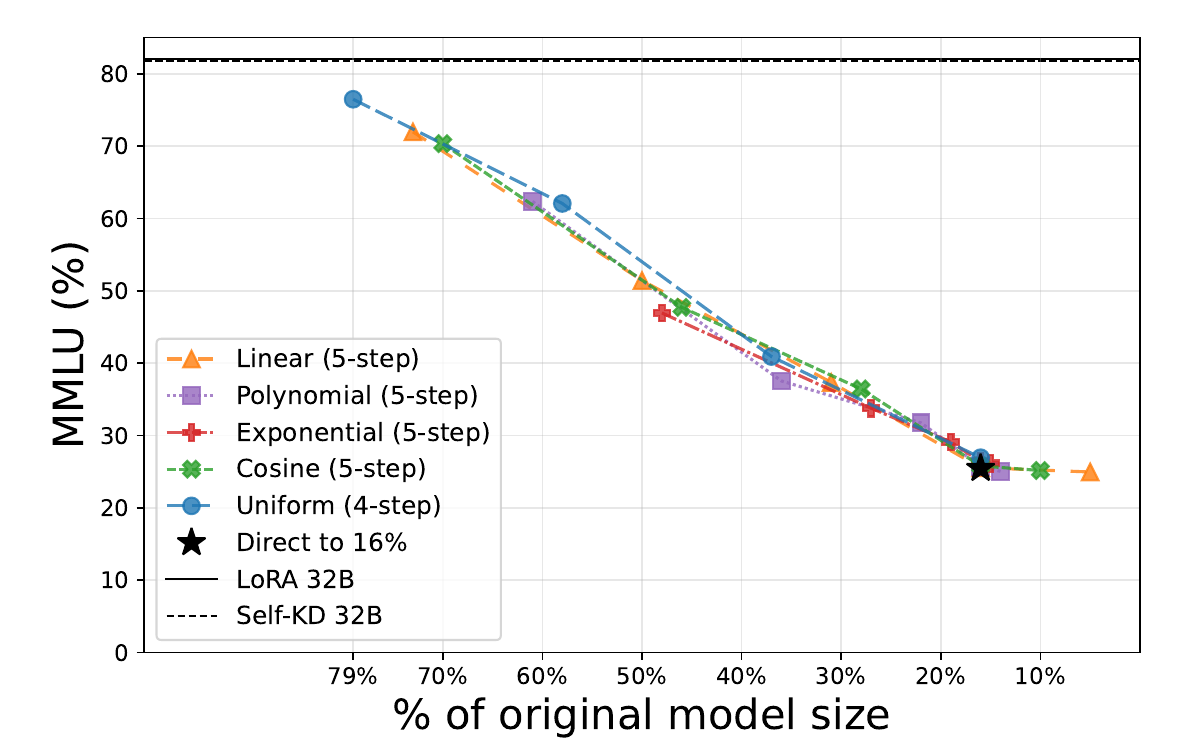}
    \smallskip
    \small (a) MMLU.
  \end{minipage}\hspace{2em}
  \begin{minipage}[t]{0.33\linewidth}
    \centering
    \includegraphics[width=\linewidth]{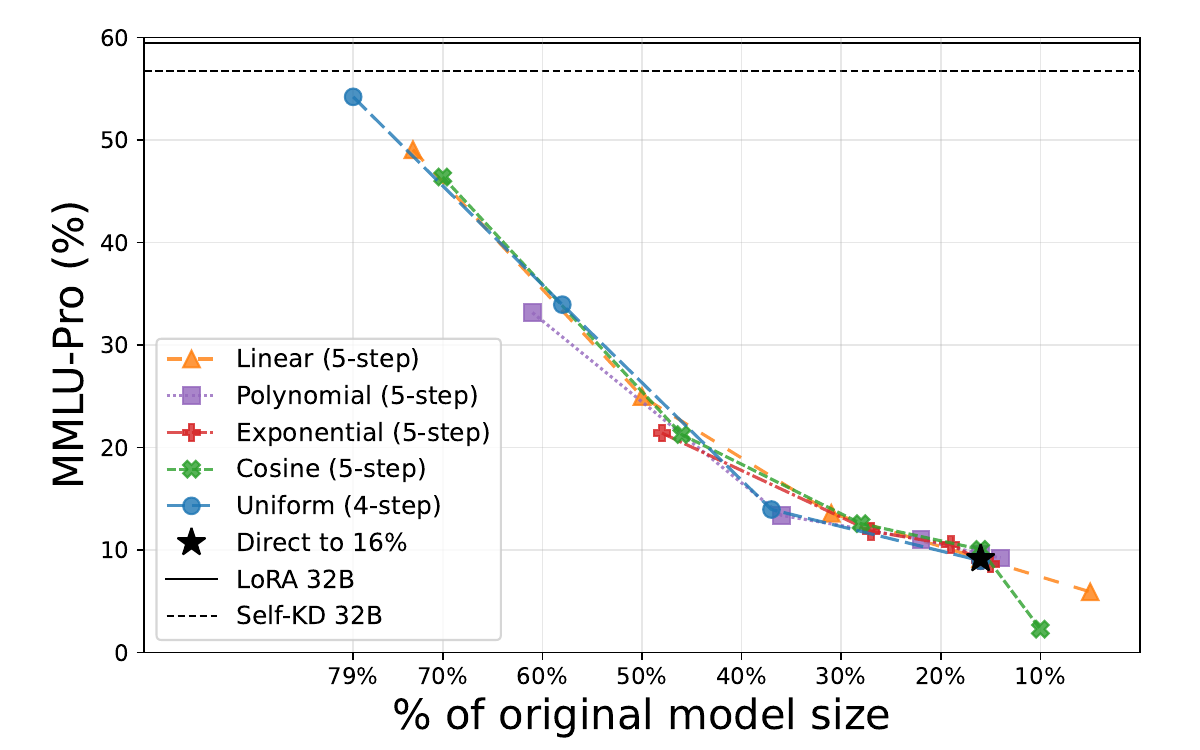}
    \smallskip
    \small (b) MMLU-Pro.
  \end{minipage}
  \caption{General-knowledge benchmarks for decayed and uniform iterative schedules: MMLU and MMLU-Pro scores at intermediate and final steps (Table~\ref{tab:schedule_comparison}). The direct-to-16\% baseline (gray star) is included as in Figure~\ref{fig:schedule_in_domain}. Same schedules, data regimes, and seeds.}
  \label{fig:schedule_mmlu}
\end{figure}

The key distinction is how aggressively each schedule front-loads the pruning: exponential decay removes 52\,pp in the first step alone, while uniform spreads the compression evenly at 21\,pp per step. In-domain results (Macro F1, NLL, Brier) and MMLU/MMLU-Pro scores at intermediate and final steps for each schedule are shown in Figures~\ref{fig:schedule_in_domain} and~\ref{fig:schedule_mmlu}.

\enlargethispage{0.8cm}

\begin{table}[!htbp]
  \enlargethispage{1cm}
  \centering
  \caption{Mean scores (across 3 seeds) at each step of the iterative compression schedules (sizes in Table~\ref{tab:schedule_comparison}). Lower is better for NLL and Brier. \textbf{Bold}: best at Step~4 (shared 16\% target). Accuracy and weighted F1 in Table~\ref{tab:sched_results_alt} (Appendix~\ref{appendix:alternative_metrics}).}
  \label{tab:sched_results}
  \scriptsize
  \renewcommand{\arraystretch}{0.95}
  \begin{tabular}{@{}l*{5}{C{2.8em}}@{}}
  \toprule
  \textbf{Schedule} & \textbf{Step 1} & \textbf{Step 2} & \textbf{Step 3} & \textbf{Step 4} & \textbf{Step 5} \\
  \midrule
  \multicolumn{6}{@{}l}{\textit{Macro F1}} \\
  \midrule
  Uniform     & 0.7724 & 0.7533 & 0.6226 & \editgreen{\textbf{0.7759}} & N/A   \\
  Linear      & 0.7738 & 0.7675 & 0.7543 & 0.7698 & 0.6429 \\
  Cosine      & 0.7752 & 0.7113 & 0.7399 & 0.7698 & 0.7447 \\
  Polynomial  & 0.7743 & 0.1041 & 0.7258 & 0.7622 & 0.7653 \\
  Exponential & 0.7480 & 0.7282 & 0.7383 & 0.7594 & 0.7550 \\
  \midrule
  \multicolumn{6}{@{}l}{\textit{Gold-label NLL $\downarrow$}} \\
  \midrule
  Uniform     & 0.6239 & 0.6575 & 0.6538 & \editgreen{0.6586} & N/A   \\
  Linear      & 0.6143 & 0.6373 & 0.6423 & \textbf{0.6471} & 0.6991 \\
  Cosine      & 0.6378 & 0.6437 & 0.6502 & 0.6566 & 0.6716 \\
  Polynomial  & 0.6175 & 0.6296 & 0.6576 & 0.6627 & 0.6742 \\
  Exponential & 0.6050 & 0.6319 & 0.6499 & 0.6600 & 0.6766 \\
  \midrule
  \multicolumn{6}{@{}l}{\textit{Brier score $\downarrow$}} \\
  \midrule
  Uniform     & 0.2980 & 0.3078 & 0.3084 & 0.3125 & N/A   \\
  Linear      & 0.2945 & 0.3033 & 0.3033 & \textbf{0.3097} & 0.3303 \\
  Cosine      & 0.3027 & 0.3086 & 0.3074 & 0.3102 & 0.3166 \\
  Polynomial  & 0.2969 & 0.3043 & 0.3114 & 0.3154 & 0.3124 \\
  Exponential & 0.2932 & 0.3064 & 0.3081 & 0.3123 & 0.3121 \\
  \midrule
  \multicolumn{6}{@{}l}{\textit{MMLU}} \\
  \midrule
  Uniform     & 76.46 & 62.06 & 40.90 & \editgreen{\textbf{26.93}} & N/A   \\
  Linear      & 71.96 & 51.42 & 37.33 & 25.54 & 24.95 \\
  Cosine      & 70.34 & 47.70 & 36.42 & 25.79 & 25.15 \\
  Polynomial  & 62.33 & 37.53 & 31.74 & 25.74 & 25.03 \\
  Exponential & 46.93 & 33.77 & 29.10 & 26.36 & 26.15 \\
  \midrule
  \multicolumn{6}{@{}l}{\textit{MMLU-Pro}} \\
  \midrule
  Uniform     & 54.21 & 33.94 & 13.95 &  \editgreen{8.95} & N/A   \\
  Linear      & 49.05 & 24.97 & 13.58 &  9.16 & 5.92 \\
  Cosine      & 46.39 & 21.26 & 12.57 & \textbf{10.06} & 2.27 \\
  Polynomial  & 33.16 & 13.38 & 11.04 &  9.61 & 9.23 \\
  Exponential & 21.41 & 11.83 & 10.52 &  8.99 & 8.63 \\
  \bottomrule
  \end{tabular}
  \end{table}

\editgreen{Table~\ref{tab:sched_results} reports mean scores (across three seeds) at each step. Figures~\ref{fig:schedule_in_domain} and~\ref{fig:schedule_mmlu} visualize the trajectories. The direct-to-16\% single-step baseline (gray star) is surpassed by all iterative schedules at the shared 16\% target. Uniform has the highest Step-4 Macro F1 (0.7759), followed closely by linear and cosine (both 0.7698), but the paths differ markedly: uniform drops to 0.6226 at Step~3 before recovering, and polynomial drops to 0.1041 at Step~2 before recovering at Step~3. These intermediate failures are transient rather than necessarily trajectory-ending.}

\editgreen{At the shared 16\% target, uniform leads Macro F1 (0.7759), while linear has the lowest NLL/Brier (NLL 0.6471 and Brier 0.3097). On MMLU, uniform is highest at 26.93, but all schedules remain near the 4-choice random-guess level. The differences between task quality, NLL/Brier, and benchmark retention reinforce that no schedule dominates every metric.}

An important observation is that intermediate-step drops in Macro F1 do not necessarily prevent subsequent recovery. Several schedules show a dip at an intermediate checkpoint where one or more seeds produce poor F1, yet the following distillation round partially or fully recovers performance. This suggests that distillation can repair some of the damage introduced by an aggressive pruning step, provided the remaining capacity is sufficient. The effect is most visible in Macro F1, which is sensitive to output formatting: if pruning disrupts the model's ability to produce correctly formatted label tokens, the argmax prediction fails even though the underlying class distribution may still be reasonable. NLL and Brier, which evaluate the full probability distribution over classes, are correspondingly more robust to such formatting failures and show smoother trajectories across intermediate steps. A focused investigation of these dips---in particular the polynomial schedule's Step~2 collapse, where Macro F1 drops to ${\approx}\,$0.10 while NLL and Brier remain in line with neighbouring schedules---is left for future work.

\editgreen{Decayed schedules have a further advantage: their diminishing late-step reductions leave budget for additional compression steps. Table~\ref{tab:sched_results} includes a Step~5 for each decayed schedule, whereas uniform cannot take a fifth 21\,pp step without crossing zero. Linear drops from 0.7698 to 0.6429 Macro F1 at 5\% model size; exponential and polynomial remain near 0.76, and cosine reaches 0.7447. MMLU is stable, but MMLU-Pro declines for linear (9.16 to 5.92) and cosine (10.06 to 2.27), so the extra step is not cost-free on every general-knowledge metric.}

\editgreen{For practitioners, the schedule choice depends on the target compression and metric: uniform has the highest Macro F1 at 16\%, linear has the lowest NLL/Brier, and exponential offers the smoothest trajectory with useful headroom beyond 16\%.}

\section{Conclusion}
\label{sec:conclusion}

\editgreen{The choice of supervision format, label-only versus blended CoT, is as consequential as the choice of compression ratio: it determines not just data efficiency but which capabilities survive compression. In the 50\% data-scaling study, blended supervision improves general-knowledge retention under both full-weight KD and LoRA. Blended CoT KD delivers the strongest in-domain classification, label-only LoRA the lowest NLL and Brier, and blended-LoRA the strongest large-data general-knowledge retention among the compressed methods, although its NLL and Brier worsen at larger data budgets.}

Self-distillation baselines on the unpruned teacher reveal that distribution shift from training alone accounts for roughly twice the gap introduced by compression to 16\%. The dominant cost is the training process itself, not the pruning.

\editgreen{Underpinning these results is the blended supervision loss, which stabilizes KL divergence distillation over CoT traces where a standard per-token loss fails. Combined with iterative pruning, it supports compression to 16\% of the teacher while maintaining meaningful task performance. The compression path matters, but severe intermediate drops can be recoverable. At the shared 16\% target, uniform achieves the highest Macro F1, linear the lowest NLL/Brier, and exponential the smoothest trajectory with headroom for further compression.}

\paragraph{Limitations.} This study uses a single dense teacher (Qwen3-32B) and one domain (financial event classification on a 35-class taxonomy). The task is short-sequence classification, and scaling behavior may differ for generative or long-context tasks. We did not explore quantization, quantization-aware distillation, cross-architecture distillation, or multi-teacher setups. Quantization is complementary to our approach and could yield additional compression gains when combined with structural pruning and distillation. Optimizer exploration was limited to AdamW with cosine annealing. Alternative optimizers may alter the stability landscape.

\paragraph{Future work.} Understanding and correcting blended-LoRA's NLL/Brier degradation at large data budgets is a natural next step; loss reweighting, rank scaling, and adaptive optimization are plausible controls. Adaptive, schedule-free optimizers are also promising for logit-based distillation: runs are short and blended CoT yields sparse, heterogeneous gradients that per-parameter learning rates could handle better than a single global rate with a fixed schedule. Validating the scaling laws on other domains, task types (generation, question answering), and teacher architectures (mixture-of-experts, hybrid) would test their generality. Combining decayed pruning schedules with quantization is a natural extension for further deployment savings. Finally, characterizing when intermediate-step dips reflect recoverable generation-formatting failures versus genuine knowledge loss---most pronounced in the polynomial schedule's Step~2 collapse (Section~\ref{sec:scheduling})---would help guide schedule design at extreme compression.

\section*{Acknowledgments}

We thank Miguel Martinez for guidance and help in the dataset creation, Guilherme Pombo for insightful early-stage discussions, and Liana Mikaelyan, Sergio Perez, Harshita Seth, Keval Morabia, and Asha Anoosheh for technical engineering support at various stages of our research.


\newpage
\appendix
{\Large\textbf{Appendix}}

\section{What Didn't Work}
\label{appendix:what_didnt_work}

We document approaches or settings that we tried but that did not improve results or proved impractical. We include this reflection to help practitioners avoid dead ends.

\begin{itemize}[leftmargin=*]
\item \textbf{Logits distillation:} A distillation temperature above 1 drastically hurts student performance. Combining the LM loss (cross-entropy) with KL divergence in a weighted sum, across several weightings we tried, was also detrimental compared to using KL divergence alone. These results are with the default AdamW optimizer and cosine annealing schedule and may not generalize to other optimizer configurations. Under cosine annealing, short warmup or an overly steep schedule (large slope between max and min learning rate) hurt performance. Long warmup and a gentle decay worked better.
\item \editgreen{\textbf{LoRA training duration:} Label-only and blended-LoRA both train stably for one epoch. Additional epochs did not improve validation scores, so we report single-epoch results. The large-data NLL/Brier tradeoff for blended-LoRA is discussed in Section~\ref{sec:data_scaling}.}
\item \textbf{Pruning allocation:} Prior work favors a more width-focused policy \cite{Muralidharan:2024aa}. We observe that heavier reliance on depth pruning degrades recovered accuracy, while overemphasizing width pruning alone induces training instability in the distilled student. Concentrating width pruning on attention heads or key/value heads in particular heavily degraded recoverable accuracy. We therefore use a balanced allocation between depth and width pruning, avoiding head-heavy or KV-heavy width pruning and combining depth with width reductions that include feedforward neurons.
\end{itemize}

\section{Homogeneous Pruning Architecture and Parameter Counts}
\label{appendix:prune}

This study is restricted to dense models. Our choice of pruning strategy is also driven by deployment: we target pruned shapes that are hardware-optimal (e.g.\ regular layer dimensions and head counts compatible with standard dense and GQA/MQA kernels). Arbitrary pruning configurations would in principle allow finer-grained compression, but would require dedicated low-level CUDA kernels per model configuration. We avoid that cost by staying within formats that existing inference stacks support efficiently. To estimate non-embedding parameter counts for decoder-only Transformer architectures, we apply a layer-wise calculation based on the dominant model components: attention projections and feedforward networks. For standard multi-head attention with a SwiGLU feedforward, the parameter count per layer is given by:
\begin{equation}
\text{Total Params} \approx N_{\text{layers}} \times \left[ 4 \times (\text{hidden size})^2 + 3 \times (\text{hidden size}) \times (\text{FFN size}) \right]
\end{equation}
where $N_{\text{layers}}$ is the number of transformer blocks, \textit{hidden size} is the embedding dimension, and \textit{FFN size} is the feedforward inner dimension. The $4 \times (\text{hidden size})^2$ term accounts for query, key, value, and output projections in the attention block, while $3 \times (\text{hidden size}) \times (\text{FFN size})$ accounts for the SwiGLU~\cite{shazeer2020glu} feedforward block (gate, up, and down projections), which usually dominates the overall parameter count.

For GQA~\cite{ainslie2023gqa} or MQA~\cite{shazeer2019fast}, where key/value projections are shared across head groups, the generalized parameter count becomes:

\begin{equation*}
\text{Total Params} \;\approx\; N_{\text{layers}} \times \bigl[\, 2\,H^2 \;+\; 2\,H^2 \cdot \tfrac{N_{\text{kv groups}}}{N_{\text{heads}}} \;+\; 3\,H \cdot F \,\bigr]
\end{equation*}
where $H$ denotes hidden size, $F$ the FFN inner dimension, $N_{\text{heads}}$ the number of attention heads, and $N_{\text{kv groups}}$ the number of distinct key/value groups. It is required that $N_{\text{kv groups}}$ divides $N_{\text{heads}}$ exactly. For standard multi-head attention, $N_{\text{kv groups}} = N_{\text{heads}}$; for GQA, $1 < N_{\text{kv groups}} < N_{\text{heads}}$; for MQA, $N_{\text{kv groups}} = 1$.

In this formulation, query and output projection parameters remain unchanged ($2H^2$), while key/value parameters are reduced by a factor of $N_{\text{kv groups}} / N_{\text{heads}}$, reflecting the efficiency gains of GQA and MQA. We use a balanced depth--width allocation. Head/KV-heavy pruning is avoided (Appendix~\ref{appendix:what_didnt_work}). The parameter-count expressions above apply to the resulting pruned architectures and are used to produce Table~\ref{tab:param_breakdown}.

\paragraph{Trainable and frozen parameters.}
\editgreen{During logit-based distillation}, only transformer block weights (attention projections and FFN weights per layer) are updated. Input and output token embeddings and the final layer normalization are frozen throughout. Table~\ref{tab:param_breakdown} gives the architecture and parameter breakdown for the compression targets used in the uniform iterative runs (Section~\ref{sec:iterative_scaling}); these cover the full size range and are representative of the broader study. A practical consequence is that frozen parameters become a progressively larger fraction of the total as the model shrinks: embeddings account for roughly 5\% of teacher parameters but around 16\% at the 16\% target, because embedding size scales with hidden dimension rather than depth. Concretely, the 16\% student has 4.42B trainable parameters, less than a seventh of the teacher's 31.21B, but retains 0.82B frozen embedding parameters, down from 1.56B in the teacher.

\begin{table}[h]
  \centering
  \caption{Architecture and parameter breakdown for the compression targets used in the uniform iterative runs (Section~\ref{sec:iterative_scaling}), which are representative of the full range studied. Trainable parameters are the transformer block weights (attention + FFN); frozen parameters are the input/output token embeddings and the final LayerNorm. Trainable \% = trainable / total.}
  \label{tab:param_breakdown}
  \small
  \begin{tabular}{@{}cccccccc@{}}
    \toprule
    \thead{Frac.\ of\\teacher size} & \thead{Layers\\(depth)} & \thead{Hidden dims\\(width)} & \thead{FFN} & \thead{Total\\(in B)} & \thead{Trainable\\(in B)} & \thead{Frozen\\(in B)} & \thead{Trainable\\\%} \\

    \midrule
    100\% & 64 & 5120 & 27904 & 32.76 & 31.21 & 1.56 & 95.2 \\
     79\% & 59 & 4736 & 25600 & 25.88 & 24.44 & 1.44 & 94.4 \\
     72\% & 57 & 4608 & 24704 & 23.59 & 22.19 & 1.40 & 94.1 \\
     58\% & 53 & 4224 & 23232 & 19.02 & 17.73 & 1.28 & 93.2 \\
     44\% & 48 & 3840 & 21056 & 14.40 & 13.24 & 1.17 & 91.9 \\
     37\% & 45 & 3648 & 19648 & 12.13 & 11.02 & 1.11 & 90.9 \\
     16\% & 33 & 2688 & 14592 &  5.24 &  4.42 & 0.82 & 84.4 \\
    \bottomrule
  \end{tabular}
\end{table}

\section{Data and Implementation Details}
\label{appendix:data}

\begin{figure}[h]
  \centering
  \includegraphics[width=0.75\linewidth]{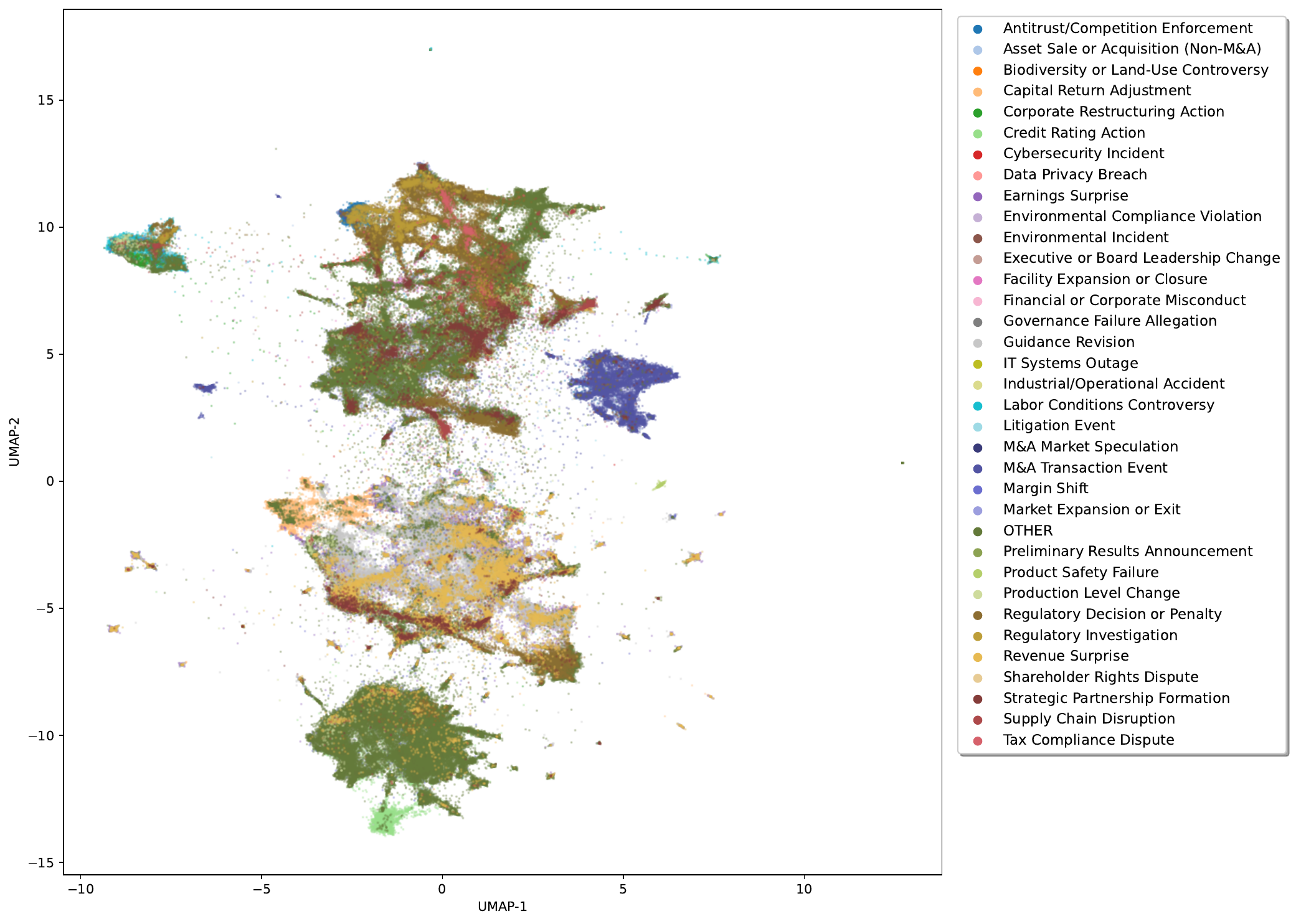}
  \caption{UMAP visualization of the dataset embeddings colored by event category. The clustering structure reflects semantic similarity between headlines within and across categories.}
  \label{fig:umap_categories}
\end{figure}

This appendix provides implementation details, including prompt formats and training configuration. The prompt was designed with subject matter expert (SME) input and refined using DSPy~\cite{Khattab:2024dspy} for prompt optimization. We use the output CoT for blended CoT distillation and the single-class label for label-only distillation, in both logit-based and LoRA-based settings. Figure~\ref{fig:umap_categories} visualizes the dataset embeddings by event category.

\subsection{Relabeling and reproducibility}
\label{appendix:teacher_eval}

We label headlines with Qwen3-32B in \editgreen{\emph{native thinking mode}} (CoT, then \texttt{[[class]]}), via vLLM~\cite{Kwon:2023vllm} with tensor parallelism ($\text{TP}=4$) and optional prefix caching, greedy at temperature~0, with incomplete runs dropped. \editgreen{The resulting corpus has ${\approx}\,$485k examples and stratified splits 400k / ${\approx}$75k / 10k (train / val / test).}

Short greedy runs without thinking (classification-only output) are bitwise repeatable across sessions. Long \editgreen{native-thinking-mode} runs can differ across sessions when inference servers change how requests are batched: kernel reductions are not necessarily \emph{batch-invariant}, so the same headline can receive numerically different activations and thus different CoT traces~\cite{He:2025nondeterminism}. To avoid conflating this inference artifact with real model error, evaluation gold labels are generated from a single greedy non-thinking session (deterministic, 100\% teacher accuracy). CoT training data are generated from a single greedy \editgreen{native-thinking-mode} session. Two independent reruns of either session agree bitwise. Training data may be relabeled in shards under different batching schedules (this affects training noise only, not the evaluation definition). \editgreen{All reported student results disable Qwen's native thinking mode; trace-trained LoRA checkpoints may nevertheless emit the CoT learned as ordinary assistant text before the final label.}

\begin{mdframed}[backgroundcolor=gray!20, linecolor=gray!50, roundcorner=2pt, linewidth=0.4pt, innerleftmargin=8pt, innerrightmargin=8pt, innertopmargin=6pt, innerbottommargin=6pt]
  \centering\small\textbf{Prompt and data format example}\par\medskip
  \hrule\medskip
  \raggedright\scriptsize
    \textbf{Instruct:} You are a helpful AI assistant that analyses financial news headlines and identifies what event type is described. You will classify event types into one of the following categories (in square brackets).

    \begin{itemize}[leftmargin=*, itemsep=1pt, parsep=0pt, topsep=3pt]
      \item \textbf{[Earnings Surprise]:} EPS beats consensus; EPS misses on costs; first profitable quarter achieved.
      \item \textbf{[Revenue Surprise]:} Subscription growth drives sales beat; demand slowdown triggers miss; revenue flat vs growth expectations.
      \item \textbf{[Margin Shift]:} Margin expands from efficiency; margin compresses from input inflation; margins stable despite shocks.
      \item \textbf{[Guidance Revision]:} Outlook raised on backlog; outlook cut after delay; guidance range narrowed on better visibility.
      \item \textbf{[Preliminary Results Announcement]:} Prelim results released early; earnings timing delayed; interim update revises prior expectations.
      \item \textbf{[Capital Return Adjustment]:} Dividend increased; buyback paused; first-ever repurchase program initiated.
      \item \textbf{[Credit Rating Action]:} Upgrade on deleveraging; downgrade on liquidity; outlook revised negative.
      \item \textbf{[Asset Sale or Acquisition (Non-M\&A)]:} Non-core unit sold; bolt-on acquisition of a small operator; minority stake disposed.
      \item \textbf{[M\&A Transaction Event]:} Deal announced; deal closes; deal terminated.
      \item \textbf{[M\&A Market Speculation]:} PE bid rumor; consolidation chatter; strategic alternatives hinted.
      \item \textbf{[Corporate Restructuring Action]:} Layoffs announced; org structure simplified; business line shut with charges.
      \item \textbf{[Market Expansion or Exit]:} New geography entered; country operations exited; new customer segment entered.
      \item \textbf{[Strategic Partnership Formation]:} Manufacturing JV created; cloud/data partnership signed; distribution alliance formed.
      \item \textbf{[Executive or Board Leadership Change]:} CEO replaced; CFO resigns; board refreshed with new independent directors.
      \item \textbf{[Governance Failure Allegation]:} Proxy adviser flags oversight; shareholder suit alleges governance failures; weak controls criticized.
      \item \textbf{[Litigation Event]:} Patent suit filed; class action settled; adverse court ruling update.
      \item \textbf{[Regulatory Investigation]:} Inquiry opened; watchdog probes complaints; investigation closed without action.
      \item \textbf{[Regulatory Decision or Penalty]:} Product approved; fine issued; approval rejected or decision delayed.
      \item \textbf{[Antitrust/Competition Enforcement]:} Merger blocked; dominance probe launched; anti-competitive fine imposed.
      \item \textbf{[Tax Compliance Dispute]:} Deductions challenged; offshore structure questioned; tax rate revised after new rules.
      \item \textbf{[Environmental Incident]:} Spill/leak; emissions exceedance; contamination event triggers cleanup.
      \item \textbf{[Environmental Compliance Violation]:} Permit breach; non-compliance notice; reporting failure cited.
      \item \textbf{[Biodiversity or Land-Use Controversy]:} Habitat impact lawsuit; deforestation backlash; protected-area land-use dispute.
      \item \textbf{[Labor Conditions Controversy]:} Discrimination claims; labor walkout; harassment allegations prompt review.
      \item \textbf{[Product Safety Failure]:} Recall launched; injuries reported; watchdog warning issued.
      \item \textbf{[Data Privacy Breach]:} Misconfigured server exposes data; consent issues probed; vendor leak impacts users.
      \item \textbf{[Financial or Corporate Misconduct]:} Accounting irregularities; bribery probe; short seller alleges inflated metrics.
      \item \textbf{[Shareholder Rights Dispute]:} Dual-class challenged; proxy fight escalates; disclosure contested before vote.
      \item \textbf{[Production Level Change]:} Capacity ramp; temporary output cuts; volume guidance lowered.
      \item \textbf{[Supply Chain Disruption]:} Port congestion delays; single-source outage; freight shock raises costs.
      \item \textbf{[Facility Expansion or Closure]:} New hub opens; plants closed; site expanded with capex increase.
      \item \textbf{[Industrial/Operational Accident]:} Explosion damages unit; derailment disrupts logistics; construction accident halts project.
      \item \textbf{[IT Systems Outage]:} Payments downtime; ERP failure delays billing; airline systems glitch disrupts ops.
      \item \textbf{[Cybersecurity Incident]:} Ransomware attack; phishing compromises accounts; unauthorized cloud access detected.
    \end{itemize}

    If the headline doesn't match any of the classes, classify it as \textbf{OTHER}.

    \textbf{ATTENTION:}
    \begin{itemize}[leftmargin=*, itemsep=1pt, parsep=0pt, topsep=2pt]
      \item Only assign a category if the headline meets all the criteria listed for that category. Otherwise use OTHER.
      \item Encourage precise matching rather than assigning categories based on partial or superficial similarities.
      \item OTHER is the default category when in doubt.
      \item If there are no specific companies mentioned, use OTHER.
    \end{itemize}

    \textbf{A few examples:}
    \begin{enumerate}[leftmargin=*, itemsep=0pt, parsep=0pt, topsep=2pt]
      \item ACME Corp Q4 EPS tops estimates as pricing power offsets softer volumes $\rightarrow$ [Earnings Surprise]
      \item Sunline beats revenue forecasts on subscription growth; guidance unchanged $\rightarrow$ [Revenue Surprise]
      \item Cobalt Industries expands operating margin after plant automation rollout $\rightarrow$ [Margin Shift]
      \item HelioTech raises full-year outlook on strong backlog and new orders $\rightarrow$ [Guidance Revision]
      \item Vantage issues preliminary Q2 results ahead of earnings; sales stronger than expected $\rightarrow$ [Preliminary Results Announcement]
      \item HarborHoldings boosts dividend 15\% and extends \$2B buyback program $\rightarrow$ [Capital Return Adjustment]
      \item S\&P upgrades Meridian to BBB- on deleveraging progress $\rightarrow$ [Credit Rating Action]
      \item IronPeak sells legacy chemicals unit for \$1.1B; plans balance-sheet reset $\rightarrow$ [Asset Sale or Acquisition (Non-M\&A)]
      \item Summit to acquire Quanta Bio in \$4.6B cash-and-stock deal $\rightarrow$ [M\&A Transaction Event]
      \item Shares jump as report says private equity weighs bid for Harbor Media $\rightarrow$ [M\&A Market Speculation]
      \item Northwind to cut 8\% of workforce as it targets \$600M in annual savings $\rightarrow$ [Corporate Restructuring Action]
      \item BrightFoods enters India through local JV; targets premium segment $\rightarrow$ [Market Expansion or Exit]
      \item Orion and Kappa form battery JV to secure supply and scale production $\rightarrow$ [Strategic Partnership Formation]
      \item Lumen names insider as CEO; founder to remain chair $\rightarrow$ [Executive or Board Leadership Change]
      \item Proxy adviser urges vote against directors over oversight concerns $\rightarrow$ [Governance Failure Allegation]
      \item Patent holder files suit against Nova Devices over flagship product $\rightarrow$ [Litigation Event]
      \item Regulator opens inquiry into Helix pricing practices, shares fall $\rightarrow$ [Regulatory Investigation]
      \item FDA approves Orchid's new therapy; launch slated for Q2 $\rightarrow$ [Regulatory Decision or Penalty]
      \item Competition authority blocks SteelCore's acquisition of rival foundry $\rightarrow$ [Antitrust/Competition Enforcement]
      \item Tax authority challenges deductions; Granite to set aside \$250M reserve $\rightarrow$ [Tax Compliance Dispute]
      \item Chemical leak at RiverChem prompts shutdown; remediation underway $\rightarrow$ [Environmental Incident]
      \item Regulators cite Summit Metals for wastewater permit violations $\rightarrow$ [Environmental Compliance Violation]
      \item Project halted after court orders review of habitat impacts $\rightarrow$ [Biodiversity or Land-Use Controversy]
      \item Workers allege discrimination; Titan says it will conduct review $\rightarrow$ [Labor Conditions Controversy]
      \item Nova recalls smart-home devices after overheating reports $\rightarrow$ [Product Safety Failure]
      \item OptiBank discloses customer data exposure from misconfigured server $\rightarrow$ [Data Privacy Breach]
      \item Audit finds irregularities; Crescent places CFO on leave $\rightarrow$ [Financial or Corporate Misconduct]
      \item Investors challenge dual-class structure; governance vote contested $\rightarrow$ [Shareholder Rights Dispute]
      \item Orion ramps production capacity by 20\% to meet demand $\rightarrow$ [Production Level Change]
      \item Port congestion delays shipments; retailer warns of stockouts $\rightarrow$ [Supply Chain Disruption]
      \item Lattice opens new distribution hub to reduce delivery times $\rightarrow$ [Facility Expansion or Closure]
      \item Explosion damages unit at Summit refinery; restart timeline uncertain $\rightarrow$ [Industrial/Operational Accident]
      \item Payment processor outage disrupts merchants; service restored overnight $\rightarrow$ [IT Systems Outage]
      \item Ransomware attack hits NorthBridge; some systems taken offline $\rightarrow$ [Cybersecurity Incident]
      \item Kopin Chairman Fan Buys 116,400 Shares @\$2.83/Share -- Form 4 $\rightarrow$ [OTHER]
    \end{enumerate}

    \medskip
    Given the following headline:\\[4pt]
    \texttt{\#\#\# START HEADLINE \#\#\#}\\[2pt]
    \texttt{<headline>}\\[2pt]
    \texttt{\#\#\# END HEADLINE \#\#\#}\\[4pt]
  What event type best classifies it? Answer only with your predicted class and give it inside double square brackets, like \texttt{[[class]]}.
\end{mdframed}

\paragraph{Training details.} For both LoRA-based and logit-based distillation, we train on completions only: input tokens are masked so that the loss is computed solely on the teacher's output. For LoRA, this is standard causal language modeling loss (cross-entropy) on output tokens. For logit-based distillation, KL divergence is computed at each output token position, with the teacher's logits serving as soft targets. All experiments use AdamW with a cosine annealing learning-rate schedule (Section~\ref{sec:methods}). Exploring optimizers that adapt learning rates from per-parameter gradient statistics rather than global schedules is left for future work.

\section{Alternative metrics}
\label{appendix:alternative_metrics}

The main text reports Macro F1, NLL, and Brier; here we define all classification metrics and give weighted F1 and accuracy for completeness.

\paragraph{Macro F1.} For $C$ classes, let $TP_c$, $FP_c$, and $FN_c$ denote the true positives, false positives, and false negatives for class $c$. Per-class precision, recall, and F1 are
\[
\mathrm{Precision}_c = \frac{TP_c}{TP_c + FP_c}, \quad
\mathrm{Recall}_c = \frac{TP_c}{TP_c + FN_c}, \quad
F1_c = \frac{2 \cdot \mathrm{Precision}_c \cdot \mathrm{Recall}_c}{\mathrm{Precision}_c + \mathrm{Recall}_c}.
\]
The macro-averaged F1 score is the unweighted mean: $F1_{\mathrm{macro}} = \frac{1}{C} \sum_{c=0}^{C-1} F1_c$. Macro F1 treats all classes equally, so it is suitable for imbalanced datasets where minority classes matter.

\paragraph{Weighted F1.} Weighted F1 averages per-class F1 weighted by support (number of true instances per class): $F1_{\mathrm{weighted}} = \sum_{c=0}^{C-1} n_c F1_c / \sum_{c=0}^{C-1} n_c$, where $n_c = TP_c + FN_c$. It is dominated by frequent classes and complements Macro F1.

\paragraph{Accuracy.} Accuracy is the fraction of correct predictions: $\mathrm{Accuracy} = (\sum_c TP_c) / N$, where $N$ is the total number of examples. It is simple to interpret but can be misleading on imbalanced datasets; we primarily report Macro F1 in the main text.

\begin{figure}[htbp]
  \centering
  \begin{minipage}[t]{0.33\linewidth}
    \centering
    \includegraphics[width=\linewidth]{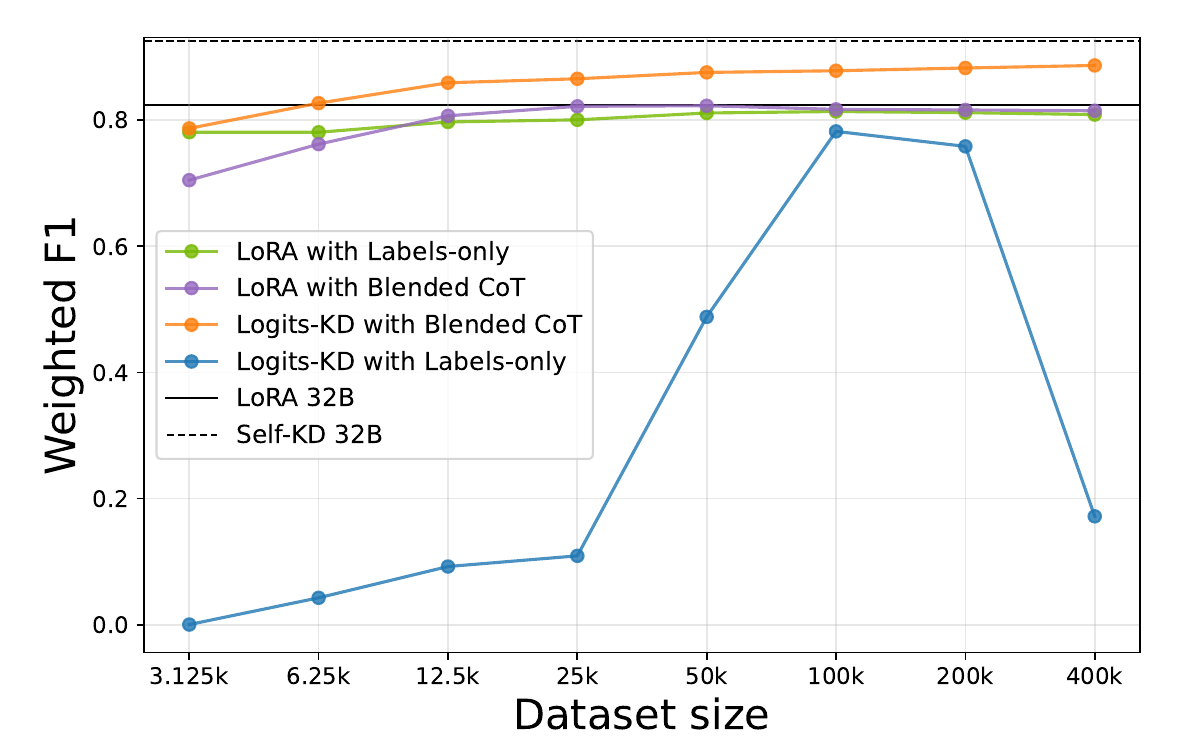}
    \smallskip
    \small (a) Weighted F1.
  \end{minipage}\hspace{2em}
  \begin{minipage}[t]{0.33\linewidth}
    \centering
    \includegraphics[width=\linewidth]{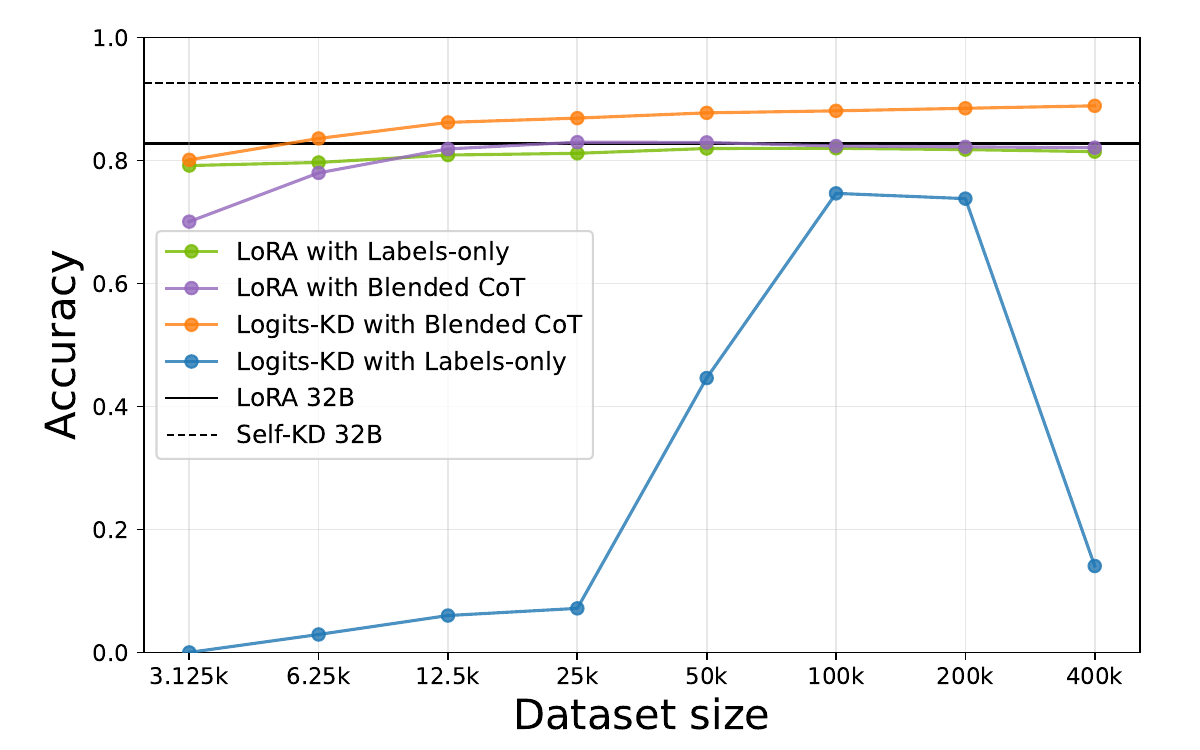}
    \smallskip
    \small (b) Accuracy.
  \end{minipage}
  \caption{In-domain classification: weighted F1 and accuracy vs.\ dataset size for the four data-scaling configurations (Section~\ref{sec:data_scaling}). Horizontal lines mark the Self-KD and LoRA baselines on the unpruned 32B model.}
  \label{fig:alt_data_scaling}
\end{figure}

\begin{figure}[htbp]
  \centering
  \begin{minipage}[t]{0.33\linewidth}
    \centering
    \includegraphics[width=\linewidth]{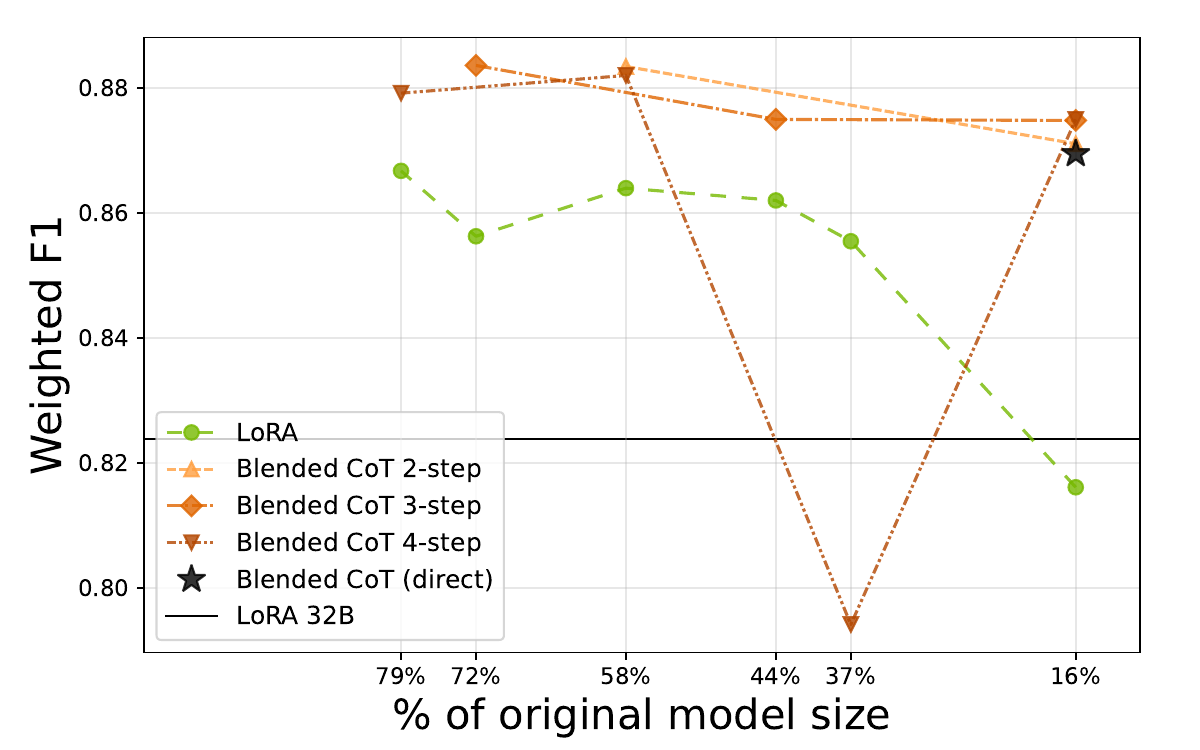}
    \smallskip
    \small (a) Weighted F1.
  \end{minipage}\hspace{2em}
  \begin{minipage}[t]{0.33\linewidth}
    \centering
    \includegraphics[width=\linewidth]{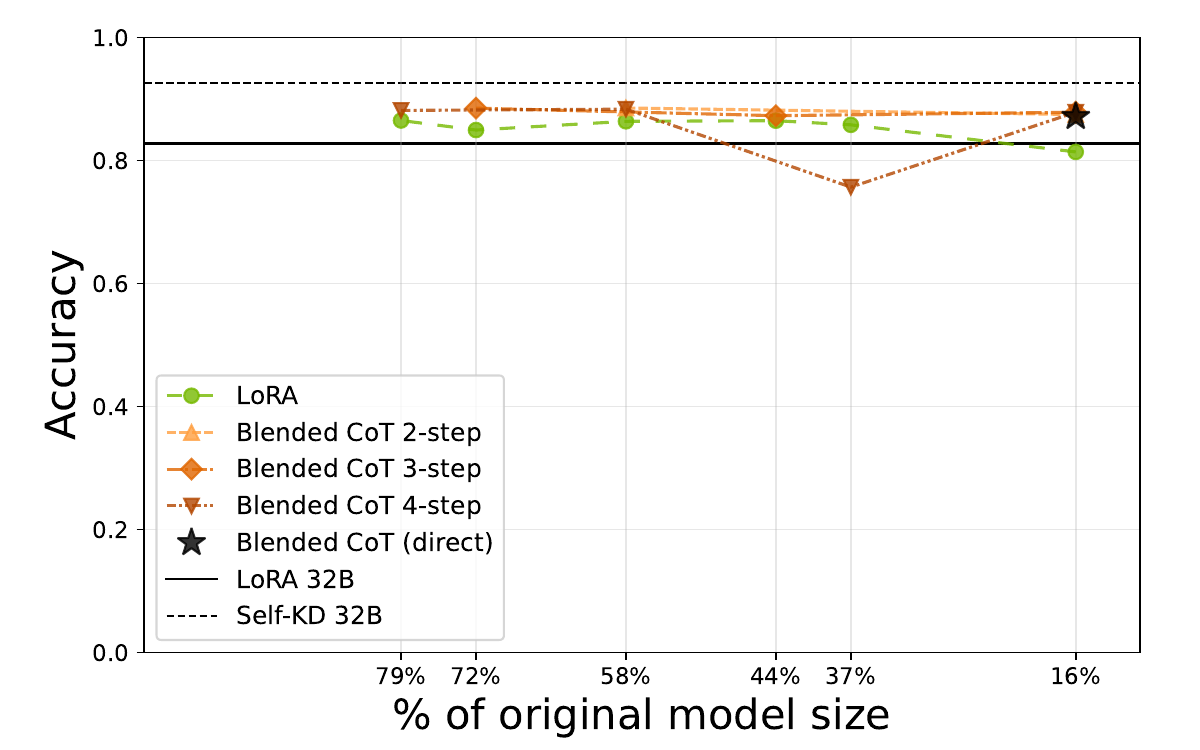}
    \smallskip
    \small (b) Accuracy.
  \end{minipage}
  \caption{In-domain classification: weighted F1 and accuracy vs.\ model size for uniform iterative scaling laws models (Section~\ref{sec:iterative_scaling}). Dotted horizontal lines mark the self-distillation baselines (Self-KD 200k and LoRA 200k, unpruned teacher).}
  \label{fig:alt_uniform_iter}
\end{figure}

\begin{figure}[htbp]
  \centering
  \begin{minipage}[t]{0.33\linewidth}
    \centering
    \includegraphics[width=\linewidth]{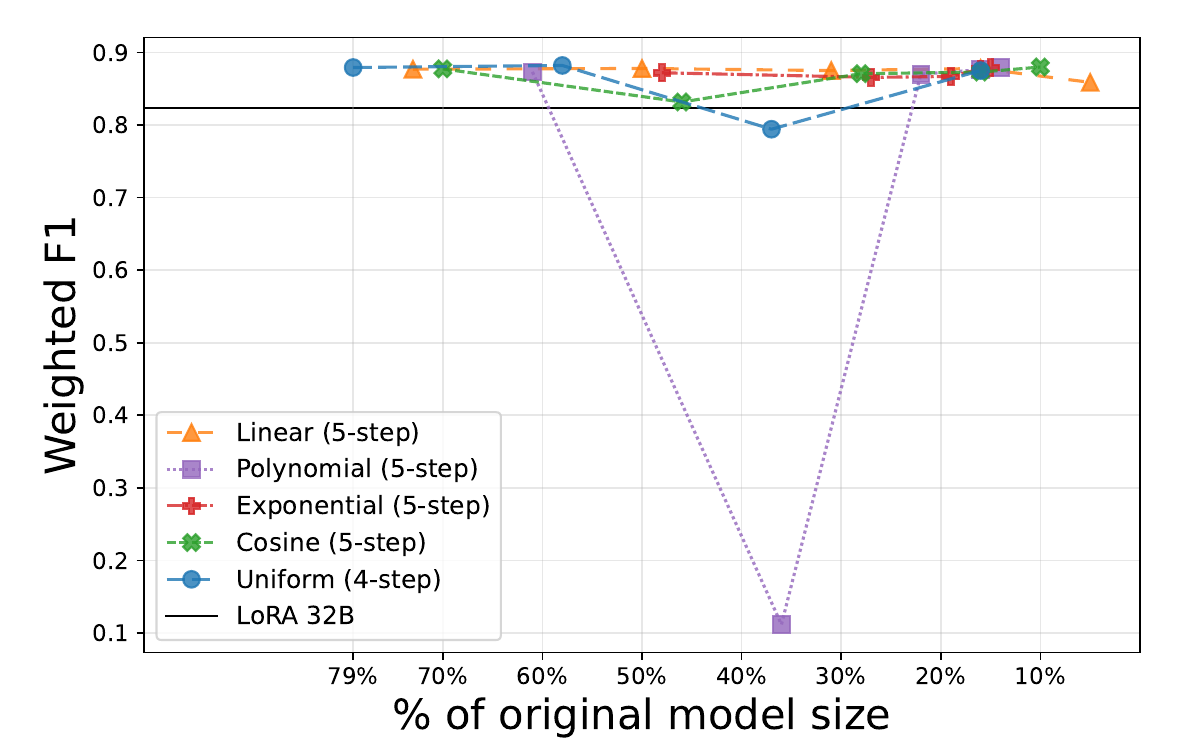}
    \smallskip
    \small (a) Weighted F1.
  \end{minipage}\hspace{2em}
  \begin{minipage}[t]{0.33\linewidth}
    \centering
    \includegraphics[width=\linewidth]{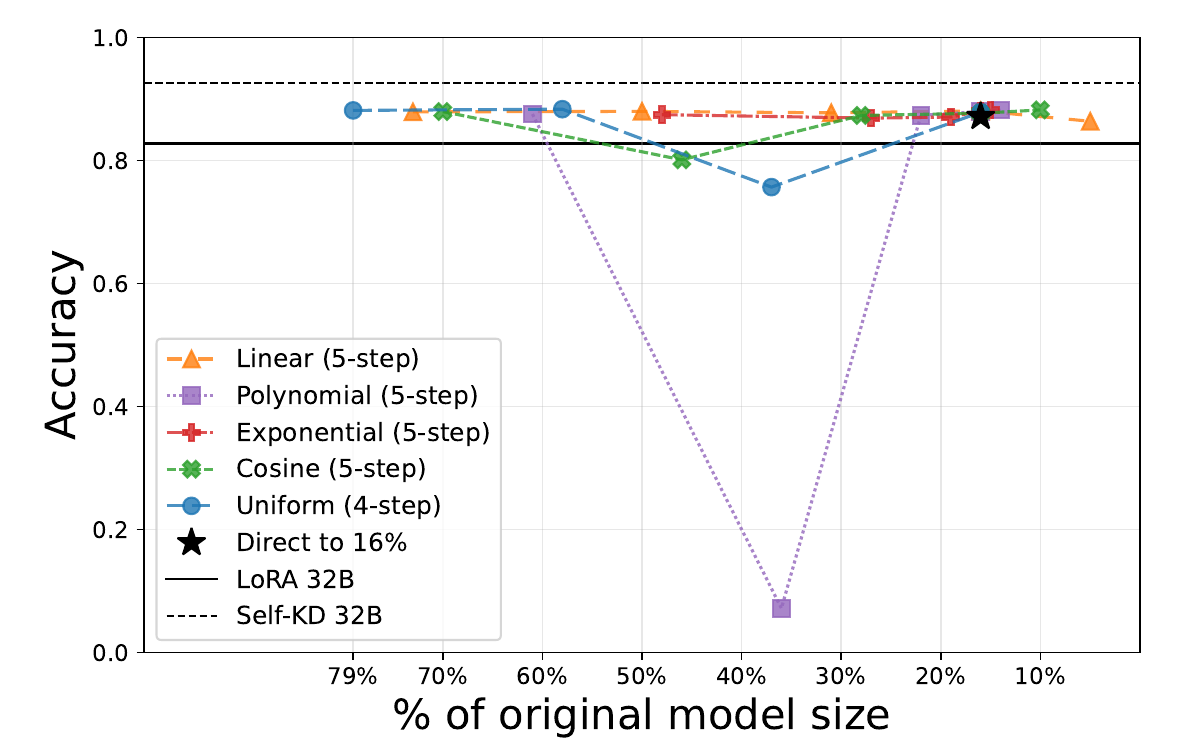}
    \smallskip
    \small (b) Accuracy.
  \end{minipage}
  \caption{In-domain classification: weighted F1 and accuracy vs.\ model size for decayed iterative scaling laws models (Section~\ref{sec:scheduling}). The dotted horizontal line marks the Self-KD 200k baseline (unpruned teacher).}
  \label{fig:alt_decayed}
\end{figure}

Figures~\ref{fig:alt_data_scaling}--\ref{fig:alt_decayed} give weighted F1 and accuracy for the data-scaling (Section~\ref{sec:data_scaling}), uniform iterative (Section~\ref{sec:iterative_scaling}), and decayed iterative (Section~\ref{sec:scheduling}) setups. Blended-LoRA peaks near 25k--50k examples (0.830 accuracy and 0.823 weighted F1) and then plateaus near 0.82 on both metrics. The most notable feature remains direct-label KD, which peaks around 50k--100k examples and then collapses at 400k. This non-monotonic failure is exposed by weighted F1 and accuracy because they are dominated by frequent classes: as training data grow, direct-label KD appears to overfit to the teacher's label distribution on majority classes while catastrophically degrading on others.

\begin{table}[htbp]
  \centering
  \caption{Accuracy and weighted F1 (mean across 3 seeds) at each step of the iterative compression schedules, complementing Table~\ref{tab:sched_results}. \textbf{Bold} marks the best value at Step~4 (the shared 16\% target).}
  \label{tab:sched_results_alt}
  \scriptsize
  \renewcommand{\arraystretch}{0.95}
  \begin{tabular}{@{}l*{5}{C{2.8em}}@{}}
  \toprule
  \textbf{Schedule} & \textbf{Step 1} & \textbf{Step 2} & \textbf{Step 3} & \textbf{Step 4} & \textbf{Step 5} \\
  \midrule
  \multicolumn{6}{@{}l}{\textit{Accuracy}} \\
  \midrule
  Uniform     & 0.8815 & 0.8834 & 0.7567 & \editgreen{0.8785} & N/A   \\
  Linear      & 0.8791 & 0.8799 & 0.8775 & \editgreen{\textbf{0.8807}} & 0.8638 \\
  Cosine      & 0.8796 & 0.8011 & 0.8732 & 0.8768 & 0.8822 \\
  Polynomial  & 0.8754 & 0.0718 & 0.8733 & 0.8796 & 0.8821 \\
  Exponential & 0.8745 & 0.8689 & 0.8707 & 0.8806 & 0.8819 \\
  \midrule
  \multicolumn{6}{@{}l}{\textit{Weighted F1}} \\
  \midrule
  Uniform     & 0.8792 & 0.8820 & 0.7942 & \editgreen{0.8750} & N/A   \\
  Linear      & 0.8769 & 0.8779 & 0.8750 & \editgreen{\textbf{0.8778}} & 0.8587 \\
  Cosine      & 0.8773 & 0.8317 & 0.8707 & 0.8729 & 0.8799 \\
  Polynomial  & 0.8732 & 0.1116 & 0.8700 & 0.8762 & 0.8792 \\
  Exponential & 0.8720 & 0.8657 & 0.8668 & 0.8773 & 0.8790 \\
  \bottomrule
  \end{tabular}
  \end{table}

\newpage

\bibliography{references}

\end{document}